\documentclass{article}
\usepackage{ijcai25}

\usepackage{times}
\usepackage{soul}
\usepackage{url}
\usepackage[hidelinks]{hyperref}
\usepackage[utf8]{inputenc}
\usepackage[small]{caption}
\usepackage{graphicx}
\usepackage{amsmath}
\usepackage{amsthm}
\usepackage{booktabs}
\usepackage{algorithm}
\usepackage{algorithmic}
\usepackage[switch]{lineno}
\usepackage{amssymb}
\usepackage{pdfpages}

\usepackage{subfigure}

\usepackage{color}
\newcommand{\rd}[1]{\textcolor{red}{#1}}
\newcommand{\bl}[1]{\textcolor{blue}{#1}}

\title{Cyclic Vision-Language Manipulator: Towards Reliable and Fine-Grained Image Interpretation for Automated  Report Generation}

\author{
    Yingying Fang$^{1}$$^{*}$, Zihao Jin$^{1}$$^{*}$, Shaojie Guo$^{2}$, Jinda Liu$^{3}$, Zhiling Yue$^{1}$, Yijian Gao$^{1}$, Junzhi Ning$^{1}$, Zhi Li$^{2}$, Simon Walsh$^{1}$, Guang Yang$^{1}$$^{\dag}$
\affiliations{
$^{1}$ {Imperial College London, London, UK}\\
$^{2}$ {East China Normal University, Shanghai, China}\\
$^{3}$ {The Chinese University of Hong Kong, Hong Kong, China}
}
\emails
\{y.fang, g.yang\}@imperial.ac.uk
}

\begin{document}

\maketitle
\setcounter{footnote}{0}  
\renewcommand{\thefootnote}{}
\footnote[1]{$^1$The appendix for this work is available at \url{https://arxiv.org/abs/2411.05261}.}

\setcounter{footnote}{0}  
\renewcommand{\thefootnote}{\arabic{footnote}}
\begin{abstract}
Despite significant advancements in automated report generation, the opaqueness of text interpretability continues to cast doubt on the reliability of the content produced.
This paper introduces a novel approach to identify specific image features in X-ray images that influence the outputs of report generation models. Specifically, we propose Cyclic Vision-Language Manipulator (\textbf{CVLM}), a module to generate a manipulated X-ray from an original X-ray and its report from a designated report generator. The essence of CVLM is that cycling manipulated X-rays to the report generator produces altered reports aligned with the alterations pre-injected into the reports for X-ray generation, achieving the term ``cyclic manipulation''.
This process allows direct comparison between original and manipulated X-rays, clarifying the critical image features driving changes in reports and enabling model users to assess the reliability of the generated texts.
Empirical evaluations demonstrate that CVLM can identify more precise and reliable features compared to existing explanation methods, significantly enhancing the transparency and applicability of AI-generated reports.
\end{abstract}

\section{Introduction}

The automated and precise interpretation of chest X-rays has transformative potential, poised to enhance the consistency, quality, and efficiency of current interpretations conducted by human experts in healthcare.
Over the past three years, substantial efforts have been invested in refining the language generation capabilities, aligning visual and linguistic features, and increasing the accuracy of clinical report findings. 
The advent of large language models has sparked another wave in report generation, prioritizing linguistic precision and sophistication \cite{lee2023llm,he2024meddr,liu2024bootstrapping}.

\begin{figure}[!ht]
    \centering
\includegraphics[width=0.95\linewidth]{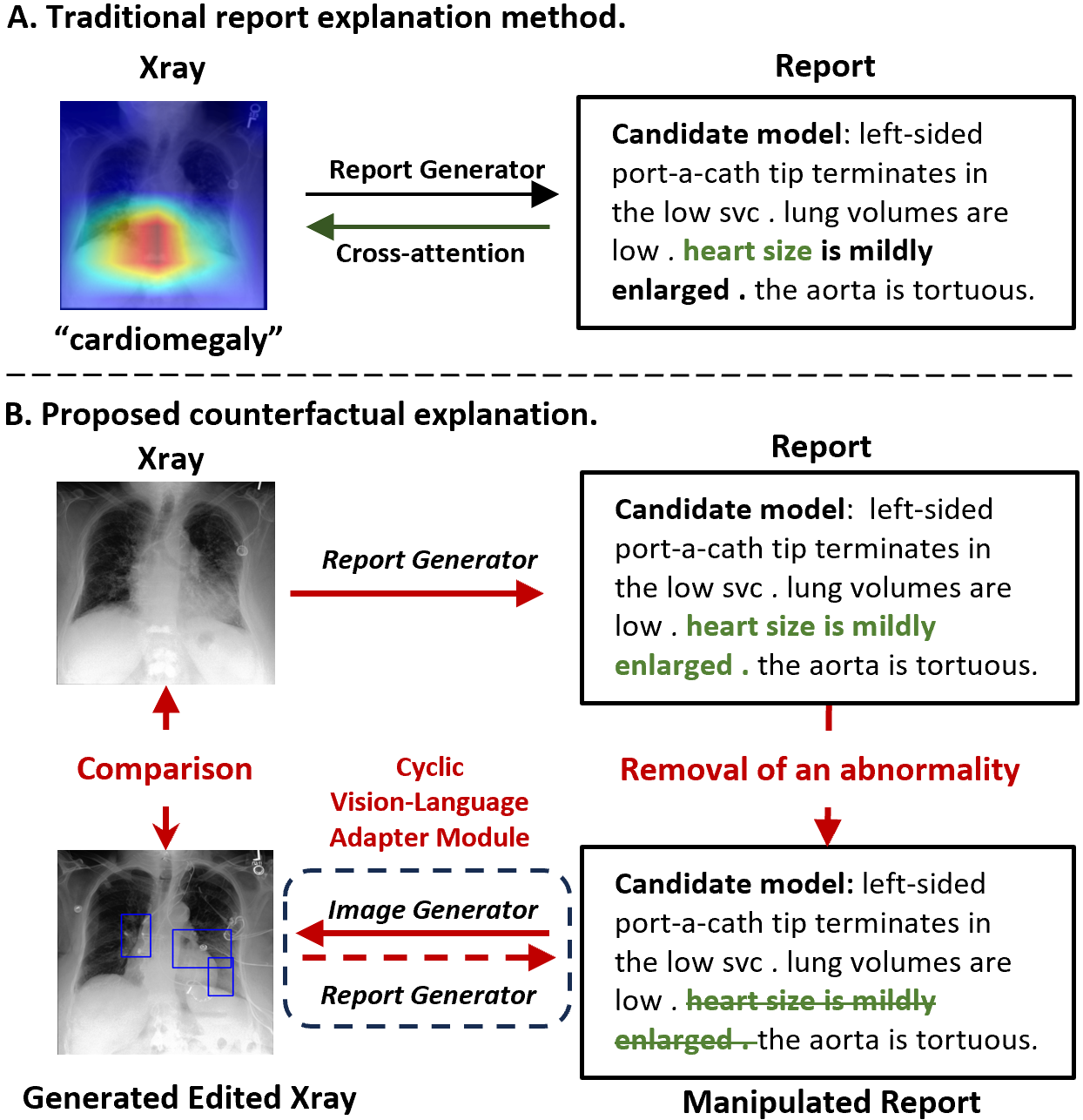}
    \caption{The overview of using counterfactual explanation for decoding the report generated from a target report generator.}
    \label{fig1}
\end{figure}
Despite these enhancements, the reports generated by these models often emerge as cryptic outputs from a ``black box'',  leaving users with little understanding of the underlying processes. Furthermore, the proliferation of diverse models leads to inconsistent reports when analyzing identical X-rays, raising concerns about the reliability of these automated systems. This variability and lack of transparency have impeded their broader adoption in clinical settings \cite{hertz2022prompt,muller2024chex}.

\begin{figure*}
    \centering
\includegraphics[width= 1\linewidth]{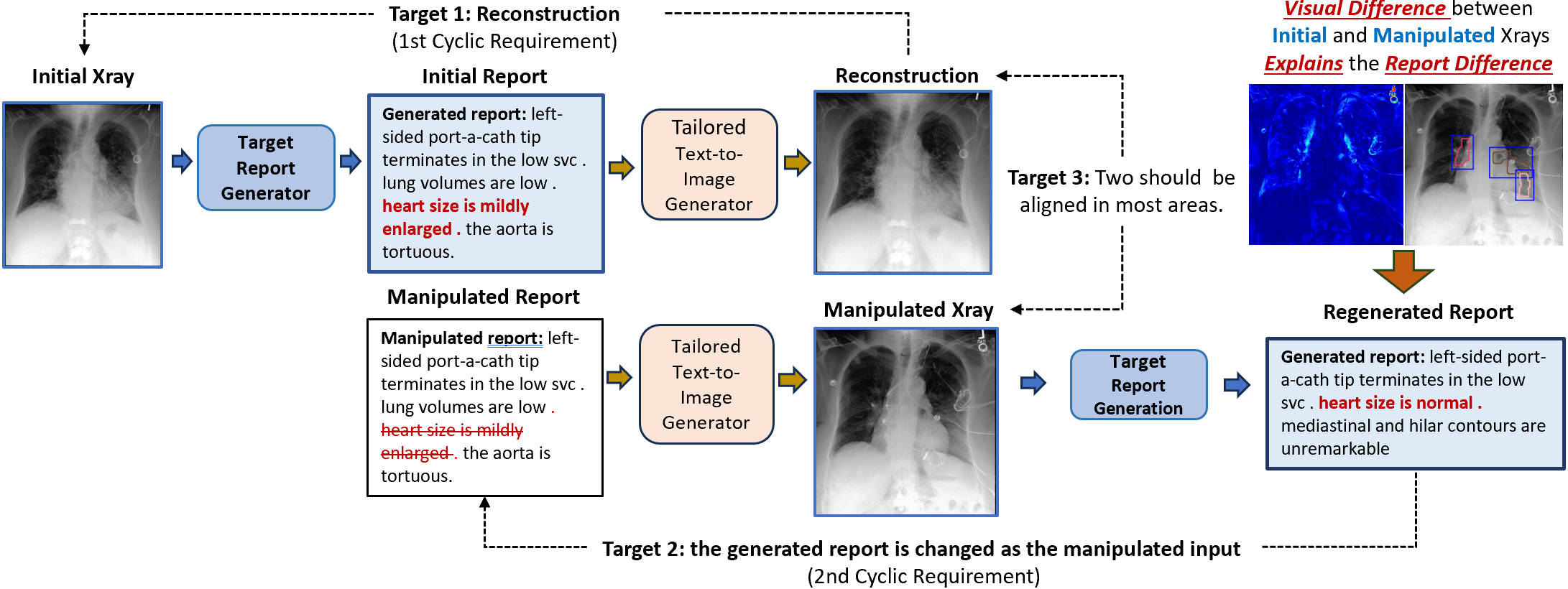}
    \caption{Overview of applying the proposed Cyclic Vision-Language Manipulator (CVLM) in explaining the report generator and the targets required for CVLM for counterfactual explanation.
    In this case, the decoded report generator identifies the abnormal heart contour as a signal for detecting ``cardiomegaly''. After reviewing the features associated with this finding, the radiologist agrees with the generated conclusion from the report generator.
}
    \label{CVLM}
\end{figure*}

In response, several studies have adopted existing Explainable AI (XAI) techniques to uncover the visual features influencing generated content, thereby aiming to enhance the trustworthiness of the generated texts. 
The most widely used XAI methods in this field typically generate an attribution map to highlight the influential features related to the generated contents through cross-attention maps \cite{liu2019clinically,cao2023mmtn,chen2020generating} or the GradCAM-based methods \cite{alfarghaly2021automated,spinks2019justifying,wang2024camanet}. 
However,  these attribution maps  often fail to highlight fine-grained visual features, instead presenting large and coarse areas (see Fig.~\ref{fig1} (A)). More critically, there is currently no reliable method to verify whether these attribution maps accurately represent the information utilized by report generators in producing findings, as there is no ground truth to confirm what the model has actually learned \cite{chase24attribution}. 
These two limitations raise an important ambiguity: while the highlighted areas in the attribution maps may appear irrelevant to the generated findings, it remains unclear whether report generators rely on irrelevant features or whether the explanation methods misidentify visual features relative to the content being explained.

To address these limitations and provide a reliable attribution map that can facilitate the understanding of generated texts, we propose a model-agnostic counterfactual explanation method to generate attribution maps. This method relies on the creation of counterfactual images, which modify specific features in the images to elicit changes in the model's decision. Comparing the counterfactual image to the original enables the identification of crucial features responsible for the altered decision \cite{wachter2017counterfactual}.
Specifically, we introduce a module called the Cyclic Vision-Language Manipulator (CVLM) to generate effective counterfactual images using a text-conditioned diffusion model. The core of CVLM is to achieve cyclic manipulation: the manipulation begins with the generated text, and the counterfactual images derived from this manipulated text must produce a regenerated report that accurately reflects the same alterations introduced in the initial report (see Fig.~\ref{CVLM} (B)).
The comparison between the cyclically generated counterfactual image and the original X-ray produces a difference map to identify fine-grained image features. Furthermore, the counterfactual images, coupled with the altered regenerated report, verify that the identified features are being utilized by existing report generators in their text generation process. 
\textbf{\textit{The major contribution of this paper}} is the development of a method capable of identifying feature attributions that can be verified as being used by the report generator for its generated content. The identified features may be consistent or inconsistent with human knowledge, as the absence of ``ground truth'' for the features learned by the model.
The verified features identified by our method can assist model developers in locating relevant visual cues used by the generator, and help human experts evaluate the reliability of the generated findings by comparing these features with their own clinical knowledge.
\textbf{\textit{The key contributions of this work}} are summarized as follows:
\begin{itemize}
\item  We propose a CVLM module for cyclic manipulation of a query X-ray and its generated reports from a designated report generator. Additionally, we introduce a \textbf{cyclic success rate} to ensure the generation of effective counterfactual images for decoding the report generation process and to quantitatively evaluate the manipulation effectiveness of CVLM. 
\item We propose an unsupervised feature localization framework based on the difference map between counterfactual and initial X-ray images, effectively filtering noise and identifying key features responsible for result changes, facilitating the understanding of explanation results without requiring additional human labeling.
\item Experiments show that counterfactual images, generated from manipulated reports, achieve fine-grained  feature identification and verification for the  generated contents from the report generator, which is unattainable with existing methods.
\item Experiments indicate that the model-agnostic nature of CVLM allows users to evaluate whether the visual features associated with the generated findings are meaningful for the corresponding content. This capability also enables users to compare the reliability of results across different report generators and make informed decisions on adopting suggestions from various models.

\end{itemize}

\section{Related Work}

\subsection{Counterfactual Explanation}

The most widely applied methods for visual explanation are post-hoc explanation methods, as their model-agnostic nature enables them to generalize across different models. Popular approaches include activation-based methods, backpropagation-based methods, and perturbation-based methods.
With the recent evolution of generative AI models, counterfactual explanation methods have emerged as an advanced perturbation-based approach. These methods generate counterfactual images that elicit different findings from the model, enabling users to identify differences between similar classes—a common challenge in medical image classification tasks. Applications has emerged in explaining the classifiers with different  modalities in medical image analysis, including X-rays \cite{atad2022chexplaining,mertes2022GANterfactual,singla2023explaining,schutte2021using,sankaranarayanan2022real,maksudov2025towards}, Magnetic Resonance Imaging \cite{tanyel2023beyond,fontanella2023acat}, ultrasound \cite{reynaud2022d}, histopathology images \cite{Karras2020ada,schutte2021using}, and Computed Tomography \cite{fang2024decoding,fang2024diffexplainer}.
Over time, counterfactual generation methods have evolved from using variational autoencoders \cite{rodriguez2021beyond} and generative adversarial networks \cite{lang2021explaining,atad2022chexplaining} to leveraging diffusion models \cite{rombach2022high,fang2024decoding,fang2024diffexplainer}.
While these methods typically generate counterfactual images to elicit changes under the guidance of a given classifier, in this work, we aim to develop a counterfactual generation method driven by text manipulation. The counterfactual X-ray is designed to achieve the desired content changes in the regenerated reports.

\subsection{Explainability in Report Generation Models}
\begin{figure*}
    \centering
\includegraphics[width=0.9\linewidth]{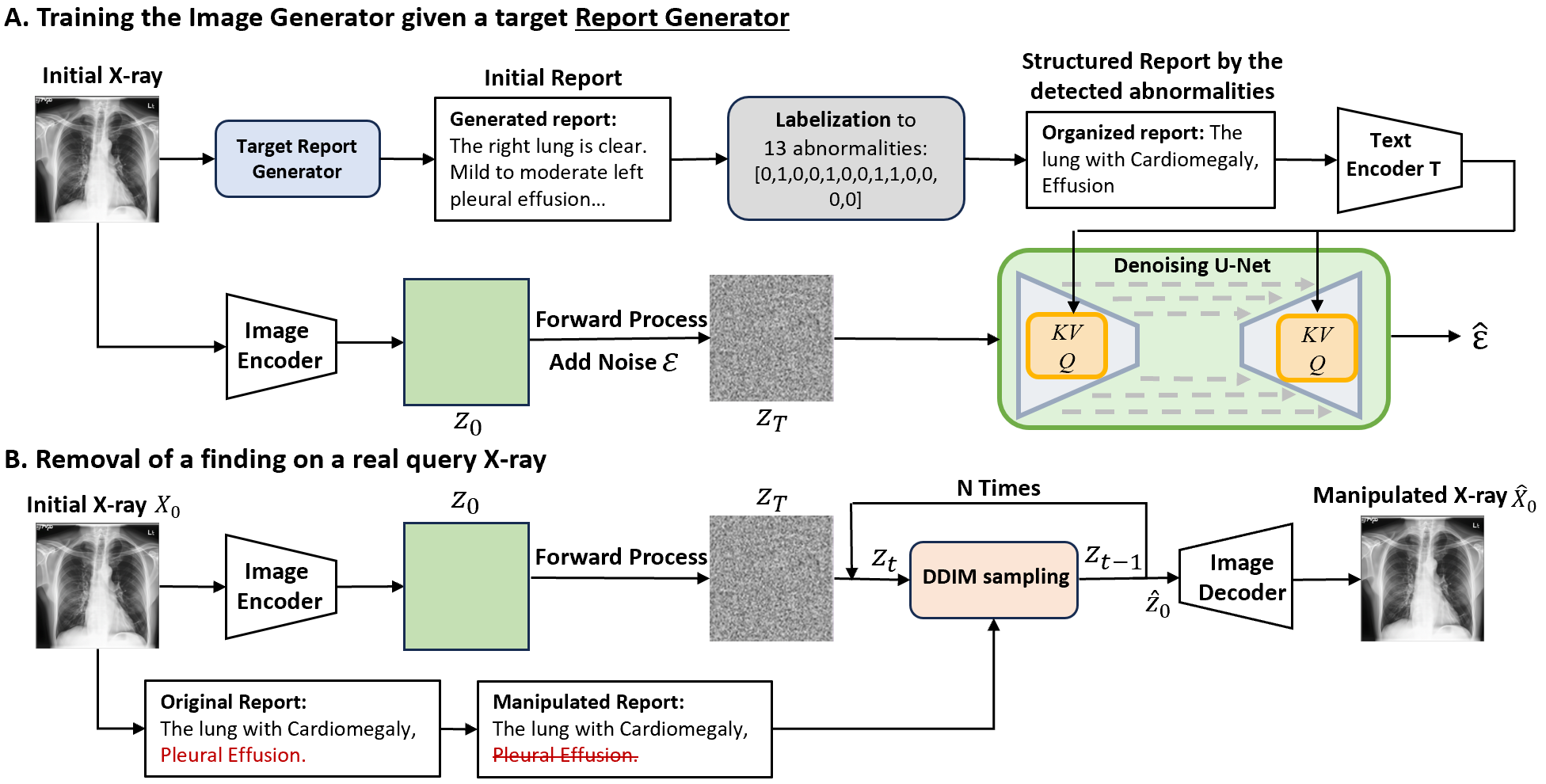}
    \caption{Overview of applying the proposed Cyclic Vision-Language Manipulator in explaining the report generator and the challenges existing in developing the CVLM.}
    \label{fig:development}
\end{figure*}

Current report generator models primarily utilize cross-attention mechanisms to generate text from input images. Consequently, most report generation methods explain their outputs by identifying the most relevant image patches for keywords in the generated report. This is achieved by calculating distances within the cross-attention architecture, which serves as the basis for the explanations of the generated keywords \cite{wang2023metransformer,cao2023mmtn,chen2023fine}.
Another category of methods \cite{alfarghaly2021automated,spinks2019justifying,wang2024camanet} employs GradCAM, a widely used technique, to identify the most activated features within the image encoder layers and produce attribution maps. However, due to their dependence on intermediate layers of deep networks, these attribution maps often face challenges such as coarse localization and ``unverifiable'' accuracy in explaining keyword-specific regions.

In contrast to these approaches, \citeauthor{tanida2023interactive}~(\citeyear{tanida2023interactive}) proposed a report generation method called RGRG, which significantly enhances the interpretability and transparency of generated reports by tailoring content generation to specific anatomical regions. However, this approach heavily relies on the preparation of a large paired dataset comprising anatomical regions and corresponding fine-grained reports for both the anatomical detection model and the report generation training. This requirement substantially increases manual labeling costs and limits the model's ability to incorporate larger X-ray-report datasets that lack fine-grained annotations. Additionally, the self-explanatory workflow of RGRG is not transferable to other report generators with different generation processes. 

In this paper, we aim to propose an explanation method that achieves fine-grained visual feature localization for generated key findings as RGRG, while eliminating the need for extensive labeling. Being model-agnostic, our approach can enhance the interpretability and transparency of generated reports across a wide range of report generation models.

\subsection{Text-controlled Image Editing}

In recent years, text-guided image editing has garnered significant interest due to its ability to simplify image editing through natural language input. A substantial body of work leverages the alignment between text and image embeddings within the pretrained large vision-language model CLIP \cite{radford2021learning}. These methods use changes in text embeddings before and after editing to infer corresponding changes in image embeddings, enabling the generation of edited images. \textit{DiffusionCLIP} \cite{kim2022diffusionclip} fine-tunes generative models using CLIP loss to align changes across the two modalities. Alternatively, other approaches map changes in text embeddings directly into the latent space of the image encoder to regenerate the image without altering network parameters \cite{patashnik2021styleclip,abdal2022clip2stylegan,lyu2023deltaedit}.

With the emergence of text-conditioned image generation models such as Stable Diffusion \cite{rombach2022high}, recent works have focused on more efficient text-guided image editing by directly modifying the text prompts used for image generation. However, a key challenge in these methods is that minor changes in prompts do not always lead to correspondingly minor changes in the generated images. To address this, \textit{Prompt-to-Prompt} \cite{hertz2022prompt} achieved localized manipulation of generated images by merging intermediate feature maps during inference. Leveraging the paired images generated by \textit{Prompt-to-Prompt}, \textit{Pix2Pix} \cite{brooks2023instructpix2pix} further introduced a more user-friendly instructive editing network. \textit{ControlNet} \cite{zhang2023adding} trains an additional control network that uses auxiliary conditions, such as sketches or segmentation maps of key objects, to keep the objects unchanged while editing other image contents via text manipulation.

While the aforementioned methods edit images based on the semantic meaning of the modified text, CVLM differs in its manipulation goal. The manipulation of counterfactual images by CVLM, driven by the manipulated text, does not need to align with real-world semantics. Instead, it is specifically designed to elicit consistent changes in the regenerated reports produced by the target report generator. By achieving cyclic manipulation, the altered images are verified to influence on the report generator, thereby uncovering whether the generator relies on clinically recognized features or other biased features.

\section{Method}

The overall framework for CVLM and applying it for decoding a target report generator is illustrated in Fig.~\ref{fig:development}. In the following, we detail the components and application of CVLM to obtain tailored explanations for a target report generator.

\subsection{Development of CVLM}

The proposed CVLM module consists of an off-the-shelf report generator that produces reports from a query X-ray and a conditional diffusion model tailored to manipulate images based on the findings detected by the report generator. To achieve the explanation capability of the generated images, the image generators in CVLM are designed to meet three specific objectives, as illustrated in Fig.~\ref{CVLM}:
(a) Reconstruction ability: Ensuring that query images can be accurately reconstructed from the unmanipulated generated reports.
(b) Minimal image change: Ensuring that altering key findings in the generated report results in only minimal changes to the image.
(c) Cyclic manipulation: Ensuring that the manipulated image results in consistent alterations in the regenerated report, completing the cyclic process.
To effectively achieve these objectives, we implemented the following adaptations based on the advanced capabilities of the text-to-image Stable Diffusion model \cite{rombach2022high}.

\subsubsection{Data Preparation}

To enable image manipulation based on the image features learned for the keywords generated by the report generator, rather than relying on real image features corresponding to medical terminology, we inferred the generated reports from the target report generator on the dataset on which it was trained. We paired the initial X-rays with the inferred results and trained the model to reconstruct the initial X-rays using the generated reports instead of the human-labeled ground-truth (GT) reports. This model is referred to as the tailored generation model, which can generate counterfactual X-rays by removing keywords from the generated reports, enabling the detection of specific differences responsible for changes in the findings produced by the target report generator.

\begin{figure}[t]
    \centering
      \subfigure[\footnotesize GT model]{
    \begin{minipage}{0.45\textwidth}
        \centering
        \makebox[0.3\linewidth]{\footnotesize{I}}
        \makebox[0.3\linewidth]{\footnotesize{R}}
        \makebox[0.3\linewidth]{\footnotesize{M}}\\
        \includegraphics[width=0.3\linewidth]{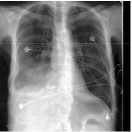} 
        \includegraphics[width=0.3\linewidth]{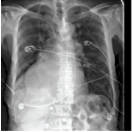} 
        \includegraphics[width=0.3\linewidth]{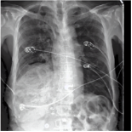}  
    \end{minipage}}\\
    \subfigure[\footnotesize Tailored model]{
    \begin{minipage}{0.45\textwidth} 
        \centering
        \makebox[0.3\linewidth]{\footnotesize{I}}
        \makebox[0.3\linewidth]{\footnotesize{R}}
        \makebox[0.3\linewidth]{\footnotesize{M}}\\
        \includegraphics[width=0.3\linewidth]{img/ablation/gd_initial.png}
        \includegraphics[width=0.3\linewidth]{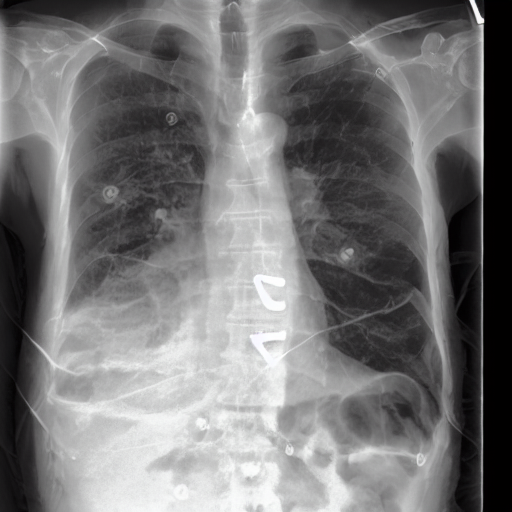} \includegraphics[width=0.3\linewidth]{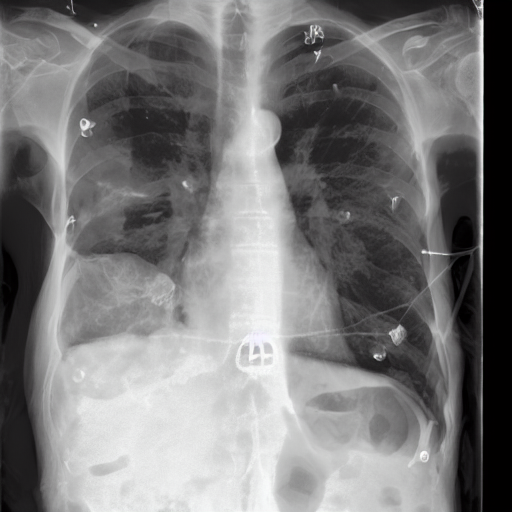}
    \end{minipage}}
    \caption{Reconstruction (R) and manipulation (M) of the initial image (I) by stable diffusion models trained with ground truth reports (GT model) and generated reports (tailored model), respectively. For both models, the reconstruction is conducted using the generated report, and the manipulation involves removing the presence of cardiomegaly from the prompt.}
    \label{motivation: training data}
\end{figure}

Conditioning the reconstruction of the query image on the generated report is also critical for further feature identification, as explained in Fig.~\ref{CVLM}. For example, as shown in Fig.~\ref{motivation: training data}, training image generators with GT reports of X-ray images  also achieve image manipulation via word manipulation. However, this approach fails to accurately reconstruct the original query X-ray. In this case, the reconstructed image from the GT model exaggerates features of ``cardiomegaly'' when it is present in the generated report but absent in the GT report. When ``cardiomegaly'' is removed from the text to observe its influence on the image, the GT reports also removes ``cardiomegaly'' but introduces more unintended changes to the initial query image. In contrast, the manipulated result from tailored model achieves more precise identification of the features in the initial image that caused the report generator to generate the finding of ``cardiomegaly''.

Furthermore, to facilitate automated manipulation of the generated reports and align the image features to the generated contents in the generated reports, we removed the redundancy and simplified the reports into a predefined list of 13 abnormalities—Enlarged Cardiomediastinum, Cardiomegaly, Lung Opacity, Lung Lesion, Edema, Consolidation, Pneumonia, Atelectasis, Pneumothorax, Effusion, Pleural Other, Fracture, and Support Devices—using the pretrained CheXbert classifier \cite{smit2020chexbert}, which is widely used for evaluating report correctness regarding major findings.
We then organized the generated reports as a concatenated list of abnormalities, formatted as: ``The lung with abnormalities ${O}_{1}$, ..., ${O}_{N}$'' where ${O}$ represents $N$ abnormalities produced in the generated reports. 

\subsubsection{Training Objective}
Our training objective follows the Stable Diffusion  training procedure, which is given as below:
\begin{equation}
L_{LDM} := \mathbb{E} \left[ \left\| \epsilon - \epsilon_\theta \left( z_T, T, \tau_\theta(y) \right) \right\|_2^2 \right], 
\end{equation}
where $z_T$ is the encoded feature of the initial real query X-ray image from the encoder of a variational autoencoder, $\tau_{\theta}$ is the text encoder that transforms the prompt to the text embedding.
During our training, we leverage the pretrained model weight for the text embedding and image autoencoder modules by a stable diffusion model pretrained on MIMIC \cite{liang2023pie}. During training, we initialize the weight of UNET architecture by the stable diffusion pretrained weight 
``CompVis/stable-diffusion-v1-4'' and freeze the parameter in the image autoencoder.

\subsubsection{Real Image Manipulation}

To enable the CVLM to explain the generated report of a real X-ray query, we employ Denoising Diffusion Implicit Models (DDIM), a non-stochastic variant of Denoising Diffusion Probabilistic Models (DDPMs), which performs the sampling process for $T$ steps:
\begin{multline}
      x_{t-1} = \sqrt{\overline{\alpha}_{t-1}} \left( \frac{x_t - \sqrt{1 - \overline{\alpha}_t} \cdot \epsilon_\theta (x_t, t)}{\sqrt{\overline{\alpha}_t}} \right)\\ + \sqrt{1 - \overline{\alpha}_{t-1} - \sigma_t^2} \cdot \epsilon_\theta (x_t, t) + \sigma_t  \epsilon_t 
      \label{ddpm}
\end{multline}
where $\epsilon_t \sim \mathcal{N}(0, \mathbf{I})$ represents a standard normal distribution, and $\sigma_t$ controls the stochasticity of the forward process. Sharing the same inference formula as DDPM, DDIM sets $\sigma_t$ in Eqn.~(\ref{ddpm}) to zero, allowing for a deterministic reconstruction without randomness. To reconstruct the initial image, we approximate the initial noise $x_T$ using DDIM Inversion, which introduces noise to the image through forward diffusion process.

\subsection{Counterfactual Explanation}

While the edited image reflects the manipulation in the report generator and achieves the desired changes in the regenerated report, as shown in Fig.~\ref{CVLM}, we refer to these manipulated X-rays as ``cyclic'' counterfactual images. These images are then utilized to decode the report generation process by identifying the visual feature changes that correspond to the modifications in the generated report for each query X-ray.

\subsubsection{Removal of Visual Abnormality}

To detect the underlying visual features associated with the context generated by the report generator, we modify the reorganized prompt by removing the findings produced in the generated report and send it to the image generation model for counterfactual generation, as shown in Fig.~\ref{fig:development} (B). A successful cyclic counterfactual image is defined as one that successfully removes the target findings in the regenerated report. We then leverage these counterfactual images to detect the visual changes that lead to the reversal of the report findings.

\subsubsection{Unsupervised Feature Identification Frame Generation}

\begin{figure}[htbp]
    \centering

    \begin{minipage}[t]{0.11\textwidth}
        \centering
        {\footnotesize Original} \\
        \includegraphics[width=\linewidth]{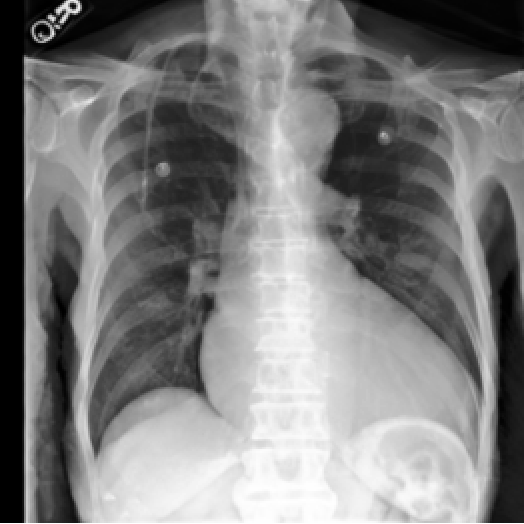}
    \end{minipage}
    \hfill
    \begin{minipage}[t]{0.11\textwidth}
        \centering
        {\footnotesize Manipulated} \\
        \includegraphics[width=\linewidth]{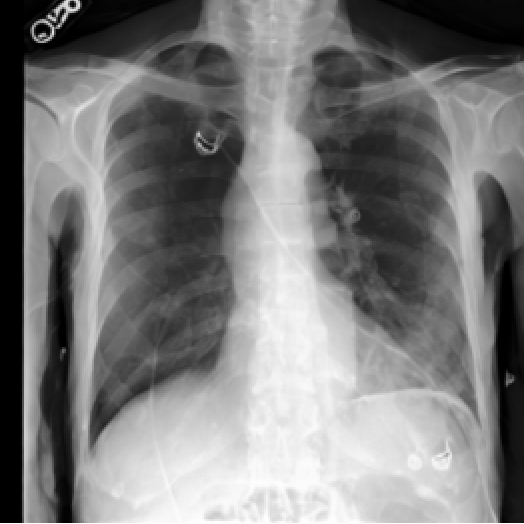}
    \end{minipage}
    \hfill
    \begin{minipage}[t]{0.11\textwidth}
        \centering
        {\footnotesize Diff (Filtered)} \\
        \includegraphics[width=\linewidth]{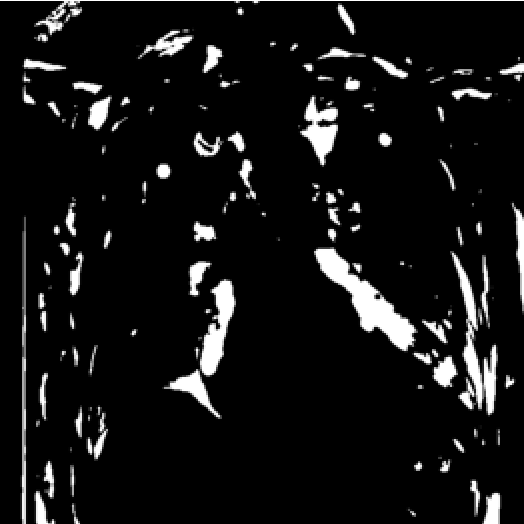}
    \end{minipage}
    \hfill
    \begin{minipage}[t]{0.11\textwidth}
        \centering
        {\footnotesize Opening} \\
        \includegraphics[width=\linewidth]{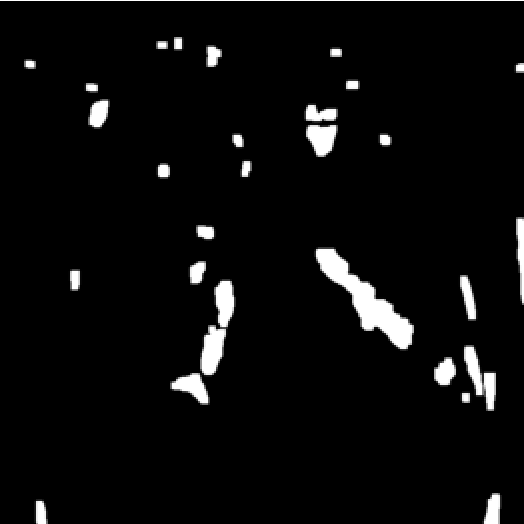}
    \end{minipage}

    \vspace{0.2cm}

    \begin{minipage}[t]{0.11\textwidth}
        \centering
        {\footnotesize Dilation} \\
        \includegraphics[width=\linewidth]{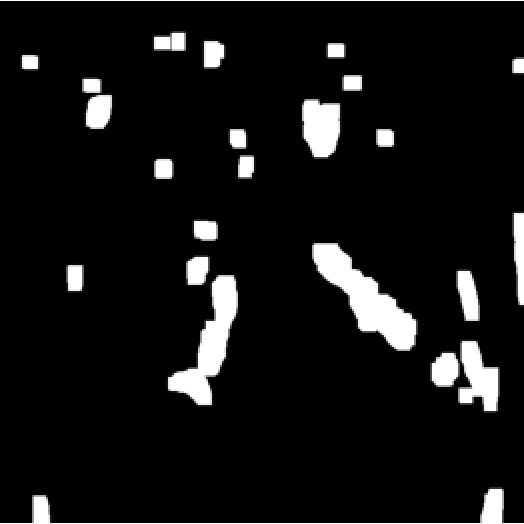}
    \end{minipage}
    \hfill
    \begin{minipage}[t]{0.11\textwidth}
        \centering
        {\footnotesize Components} \\
        \includegraphics[width=\linewidth]{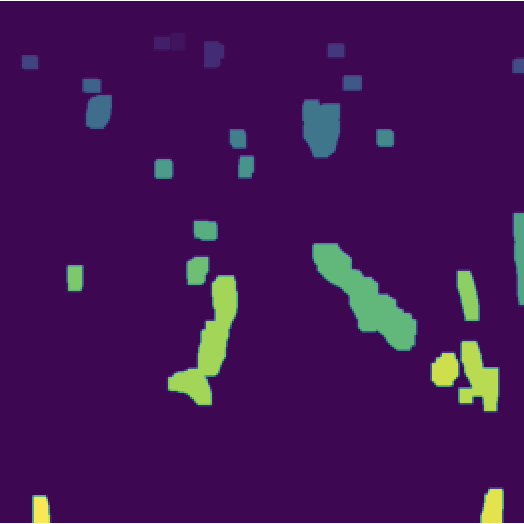}
    \end{minipage}
    \hfill
    \begin{minipage}[t]{0.11\textwidth}
        \centering
        {\footnotesize Top Contour} \\
        \includegraphics[width=\linewidth]{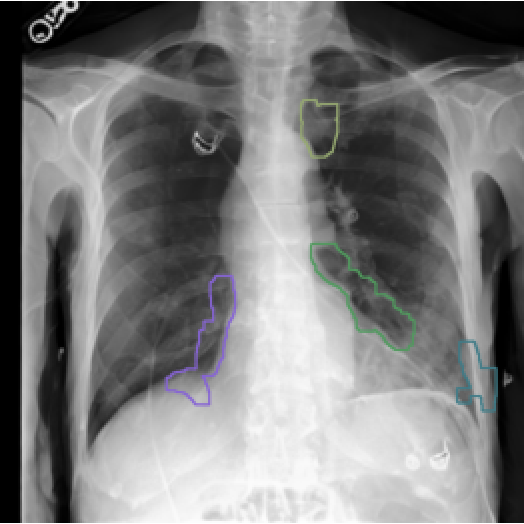}
    \end{minipage}
    \hfill
    \begin{minipage}[t]{0.11\textwidth}
        \centering
        {\footnotesize Frame} \\
        \includegraphics[width=\linewidth]{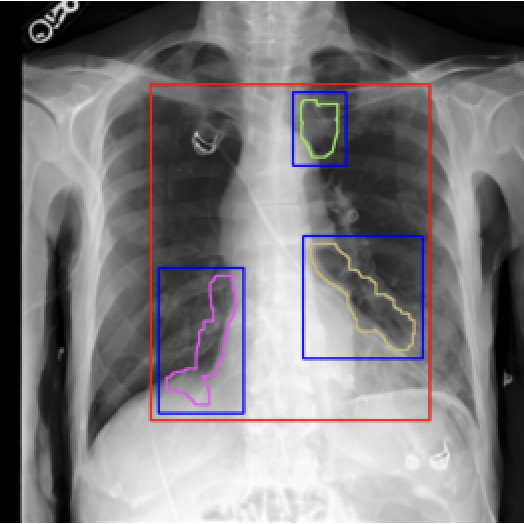}
    \end{minipage}

    \caption{Image processing pipeline in chest X-ray analysis.}
    \label{fig:xray_pipeline}
\end{figure}

To facilitate the detection of crucial features that alter the findings in the regenerated report, we propose an unsupervised method for generating feature identification frames. This method calculates a frame based on the absolute difference map between the initial X-ray and its counterfactual, allowing the observation of visual alterations responsible for changes in the report while minimizing noise and excluding isolated points.
Specifically, the absolute difference between the two images is first computed, followed by the application of a Gaussian blur with a kernel size of $H \times W$ and a threshold $L$ to suppress noise in the difference map. To identify abnormalities that are semantically represented in the image, we extract the contours of isolated pixels, group them into connected components, and retain only the most significant ones by selecting contours with the largest areas. Finally, the difference frame is constructed by retaining the top $K$ major components. An example of the entire workflow is illustrated in Fig.~\ref{fig:xray_pipeline}.

\begin{figure}[ht]
  \centering
  \subfigure[\footnotesize R2GenCMN: Cardiomegaly, Support Devices, \bl{Atelectasis}]{
      \begin{minipage}[t]{0.09\textwidth}
        \centering
        {\footnotesize Query} \\
        \includegraphics[width=\linewidth]{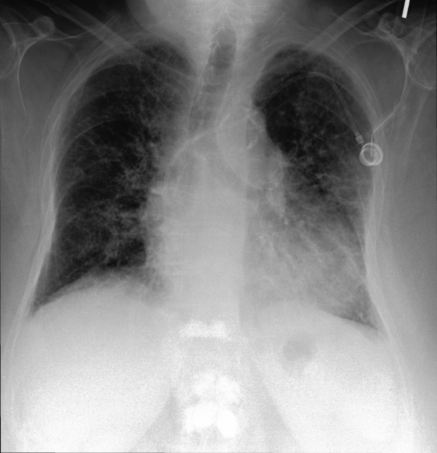}
    \end{minipage}
    \hfill
    \begin{minipage}[t]{0.09\textwidth}
        \centering
        {\footnotesize Recon} \\
        \includegraphics[width=\linewidth]{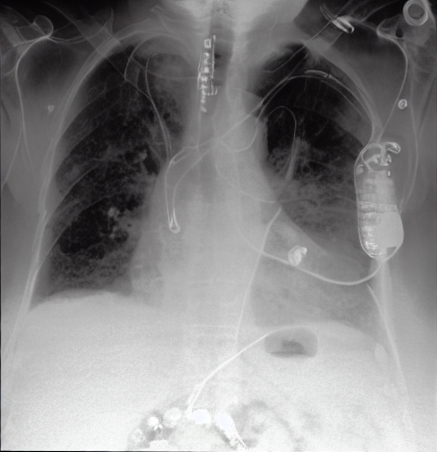}
    \end{minipage}
    \hfill
    \begin{minipage}[t]{0.09\textwidth}
        \centering
        {\footnotesize Cardiomegaly} \\
        \includegraphics[width=\linewidth]{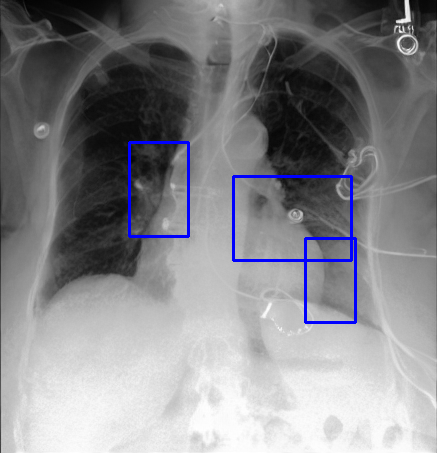}
    \end{minipage}
    \hfill
    \begin{minipage}[t]{0.09\textwidth}
        \centering
        {\footnotesize Device} \\
        \includegraphics[width=\linewidth]{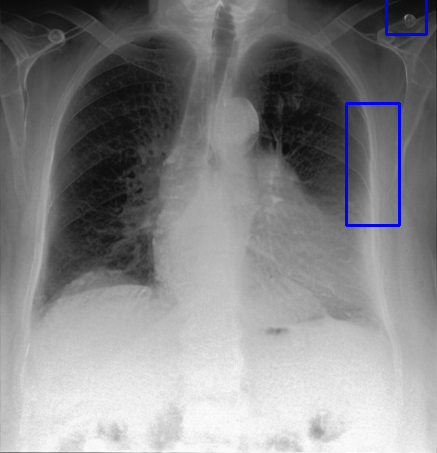}
    \end{minipage}
    \hfill
        \begin{minipage}[t]{0.09\textwidth}
        \centering
        {\footnotesize Atelectasis} \\
        \includegraphics[width=\linewidth]{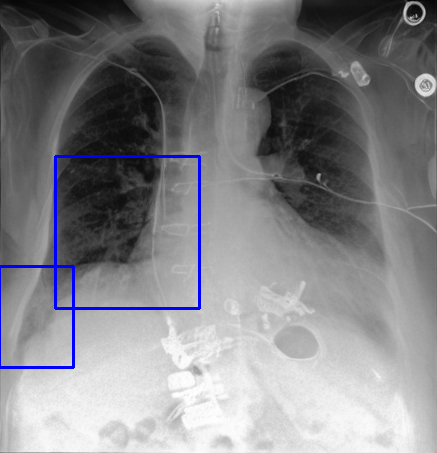}
    \end{minipage}
    }
  \subfigure[\footnotesize R2Gen: Enlarged Cardiomediastinum, Support Devices]{
    \centering
    \includegraphics[width=\linewidth]{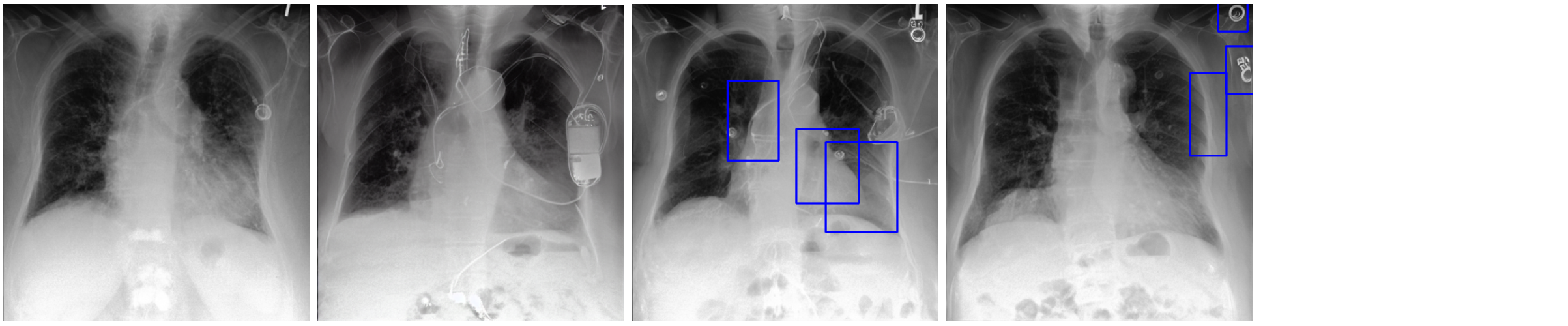}%
    \label{fig:figure7}
    }

  \caption{Explanation results for the same query X-rays with two different report generators. “(0)” denotes that the finding has been removed in the regenerated report; blue text indicates false positives against the ground-truth report.}
  \label{Figure: differentmodels}
\end{figure}

\section{Experiments}

In this section, we first outline the experimental settings, followed by the presentation of results. These results include the effectiveness of CVLM in explaining report contents, comparisons of explanation approaches, and ablation studies.

\subsection{Experimental Setting}

\subsubsection{Dataset and Report generators}
In this paper, we developed and evaluated CVLM to decode two report generators, R2Gen \cite{chen2020generating} and R2GenCMN \cite{chen2022cross}, in order to detect the visual features utilized by each within X-ray images. For each CVLM, we prepared the training dataset using MIMIC-CXR \cite{johnson2019mimic}, as it was also used to train both report generators.
The dataset comprises 473,057 chest X-ray images and 206,563 paired reports from 63,478 patients. Following the methodology in the two prior works, we utilized a subset of the dataset, consisting of 270,790 X-rays, to train CVLM. A validation set of 2,130 X-rays was used to select the optimal image generation model, while a test set of 3,858 images and reports was used to generate the corresponding counterfactual images.

\subsubsection{Implementation Details} 
For developing CVLM, we froze the text encoders and trained the diffusion model using a batch size of 8 and a learning rate of $5 \times 10^{-5}$ on a single A6000 GPU with 40 GB of memory. The model was trained for 100k steps over approximately one week. The final model for cyclic counterfactual generation was selected based on the highest PSNR of the reconstructed images on the validation set. For counterfactual generation, the DDIM step was set to 50. For frame mask generation, a Gaussian blur of size $5\times5$ was applied, and the threshold was set to $95\pm10$, with the best value selected for each manipulated finding. The number of preserved tokens, $K$, was fixed at 5.
More details are provided in the online appendix\footnotemark[1].
\footnotetext[1]{The appendix for this work is available at \url{https://arxiv.org/abs/2411.05261}.}

\subsubsection{Evaluation Methods}

We conduct a thorough evaluation of the proposed method in the following four steps:
(1) We first propose a metric called \textbf{\textit{Cyclic Rate}} \textbf{(CR)} to quantitatively assess the effectiveness of CVLM in generating cyclic counterfactual images for explanation. CR is defined as the success rate of achieving the intended manipulation in the regenerated report. Specifically, CR is calculated as the ratio of counterfactual images generated from reports where a finding is removed, and whose regenerated report successfully reflects the removal of that finding, to the total number of manipulations performed;  
(2) Next, we apply the feature identification framework to decode the visual features utilized by two individual report generators and compare the features they utilize;
(3) To further illustrate the superiority of CVLM in identifying fine-grained and verifiable features for generated contents, we compare our proposed difference detection frame (introduced in Section 3.2) with a widely used model-agnosic explanation method Cross-Attention Map and the explanation results from a state-of-the-art self-explainable report generator RGRG \cite{tanida2023interactive}. The results are shown in Fig.~\ref{Figure: differentmodels}. 
(4) Finally, we conduct ablation studies on the Cyclic Rate of CVLM, analyzing its performance with respect to inference times ($T$), training durations, and the use of raw reports versus structured reports.

\subsection{Results}

\subsubsection{Success Rate of Cyclic Counterfactual Generation}

In Table~\ref{Table: Ablation_gd}, we first investigate the impact of different training durations in the cyclic rate (CR) schedule of CVLM on generating cyclic counterfactual explanations. 
We then compare the CR settings for R2Gen and R2GenCMN.
Specifically, we evaluate models that achieve the highest reconstruction quality (measured by PSNR) and those that achieve the lowest FID score.
Our results show that the model with the best reconstruction ability, when paired with the generated text, leads to the highest cyclic manipulation effectiveness in explaining the reports for both R2Gen and R2GenCMN.
We therefore select this checkpoint for report explanation.
Both models achieve a success rate around 0.7, with CVLM for R2Gen showing a higher manipulation success rate overall.

\begin{table}[]
\centering
\footnotesize
\resizebox{0.45\textwidth}{!}{
\begin{tabular}{llll}
\toprule
Remove\_success  & GT    & Model\_16k & Model\_46k \\
\midrule
R2GenCMN & 0.655 & 0.690        & 0.595 \\
\midrule
Remove\_success  & GT    & Model\_14k & Model\_42k \\
\midrule
R2Gen      & 0.703 & 0.712        & 0.665 \\
\bottomrule
\end{tabular}}
\caption{Success rate of CVLM in explaining R2GenCMN and R2Gen across training checkpoints selected by PSNR and FID. Success rates are calculated over 569 findings from 400 images. 14k and 16k denote best-PSNR checkpoints; other iterations are selected based on lowest FID.}
\label{Table: Ablation_gd}
\end{table}

\begin{figure*}[ht]
    \centering
    \resizebox{0.95\textwidth}{!}{
    \begin{tabular}{cccccc}
        \begin{minipage}{0.15\textwidth}
            \makebox[\linewidth]{\centering \large Support Device} \\
            \includegraphics[width=\linewidth]{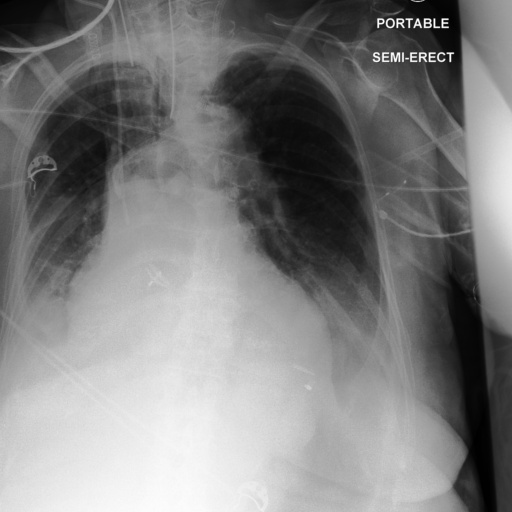} \\
            \includegraphics[width=\linewidth]{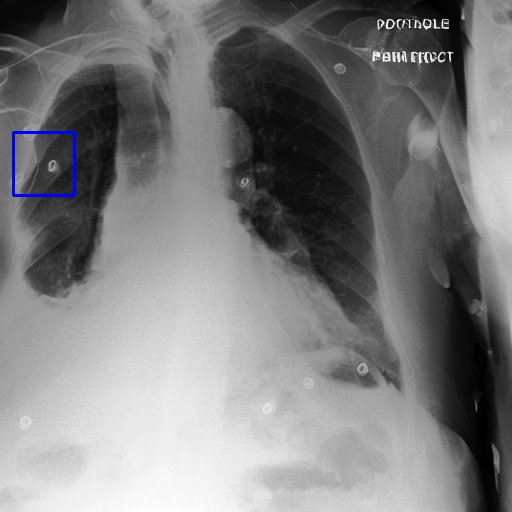} \\
            \includegraphics[width=\linewidth]{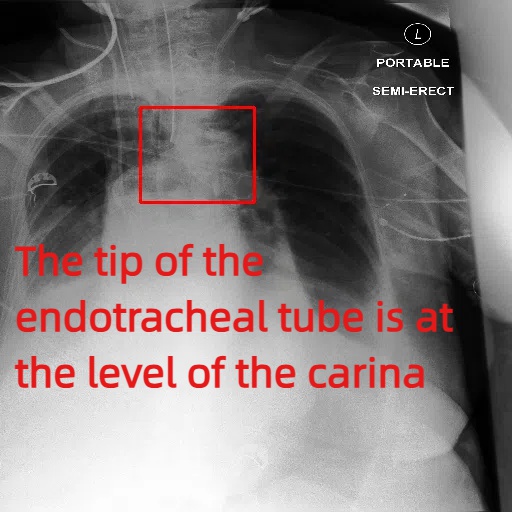} \\
            \includegraphics[width=\linewidth]{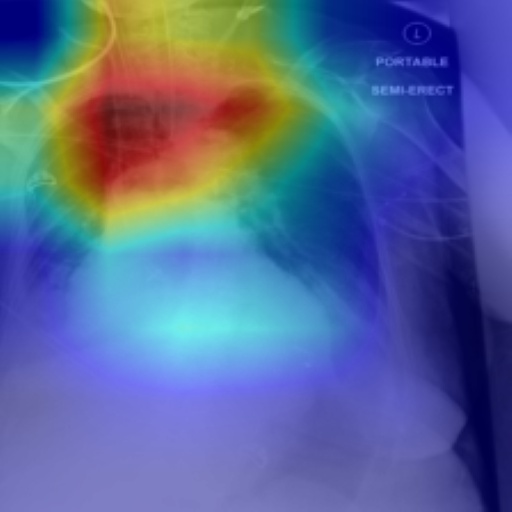} \\
            \makebox[\linewidth]{\centering \large \bl{`endotracheal tube'}}
        \end{minipage} &
        \begin{minipage}{0.15\textwidth}
            \makebox[\linewidth]{\centering \large Cardiomegaly} \\
            \includegraphics[width=\linewidth]{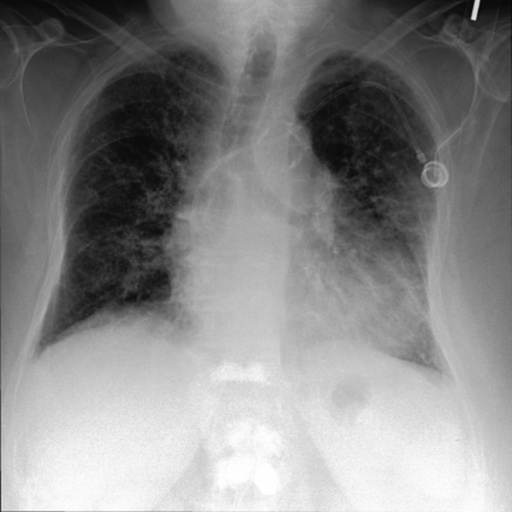} \\
            \includegraphics[width=\linewidth]{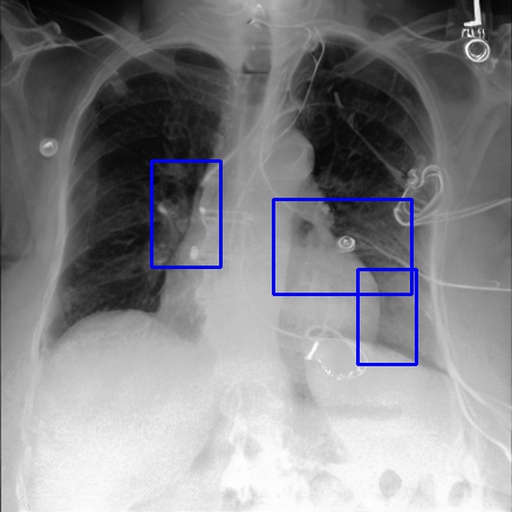} \\
            \includegraphics[width=\linewidth]{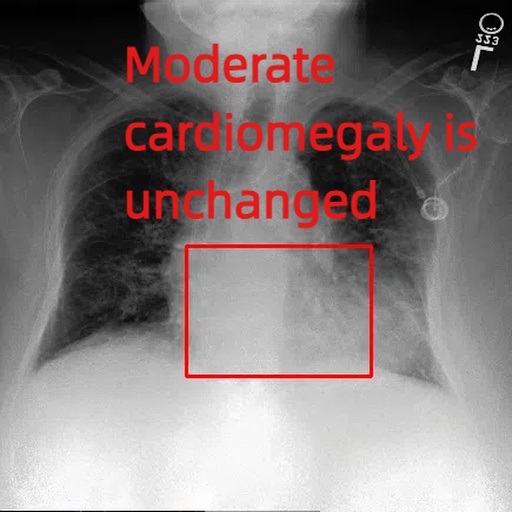} \\
            \includegraphics[width=\linewidth]{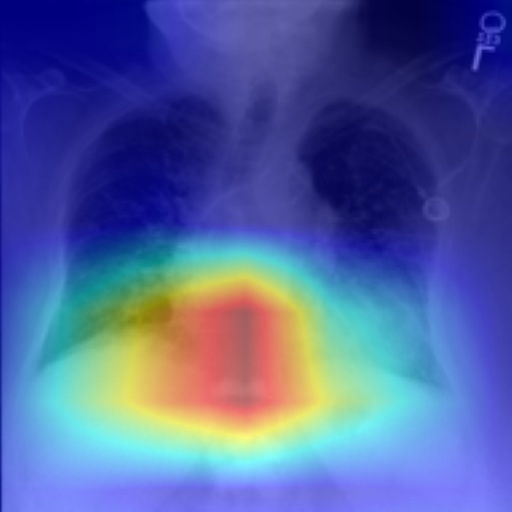} \\
            \makebox[\linewidth]{\centering \large \bl{`heart size'}}
        \end{minipage} &
        \begin{minipage}{0.15\textwidth}
            \makebox[\linewidth]{\centering \large Cardiomediastinum} \\
            \includegraphics[width=\linewidth]{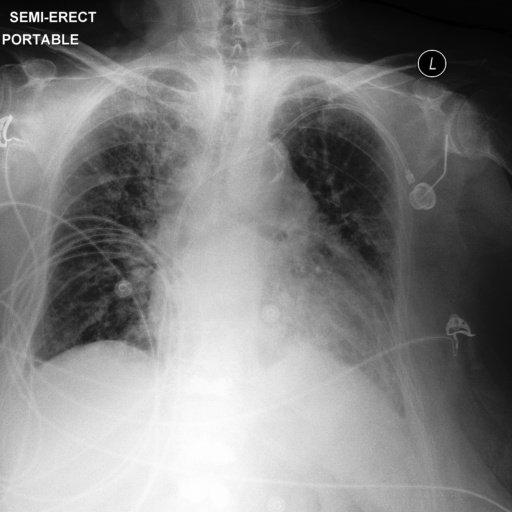} \\
            \includegraphics[width=\linewidth]{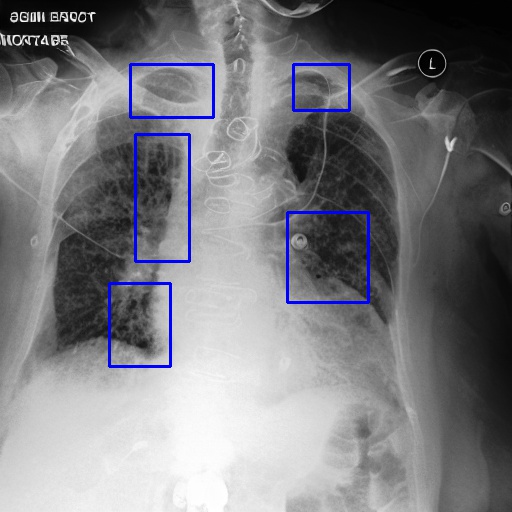} \\
            \includegraphics[width=\linewidth]{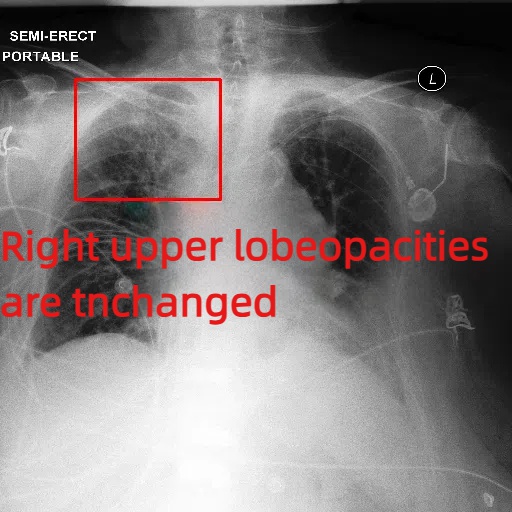} \\
            \includegraphics[width=\linewidth]{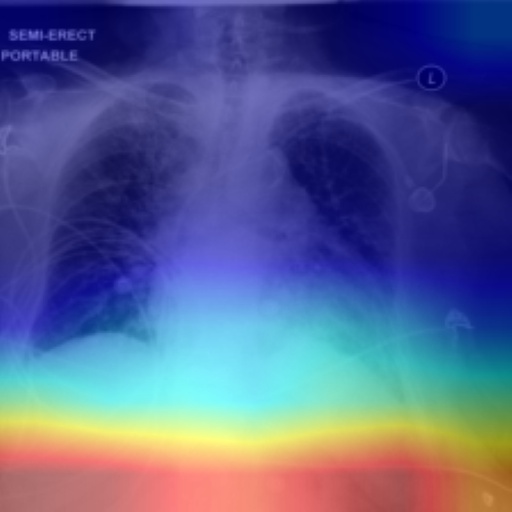} \\
            \makebox[\linewidth]{\centering \large \bl{`cardiacsilhouette'}}
        \end{minipage} &
        \begin{minipage}{0.15\textwidth}
            \makebox[\linewidth]{\centering \large Edema} \\
            \includegraphics[width=\linewidth]{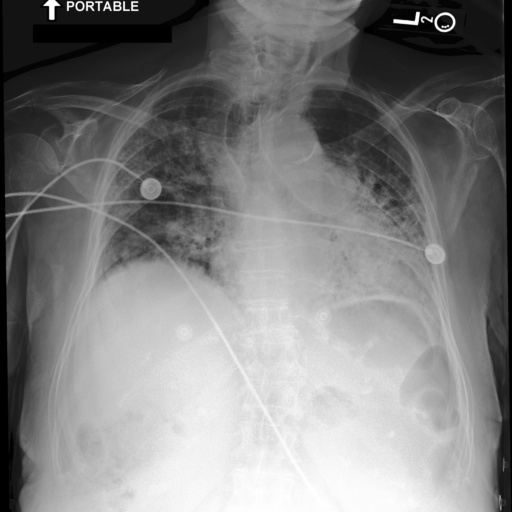} \\
            \includegraphics[width=\linewidth]{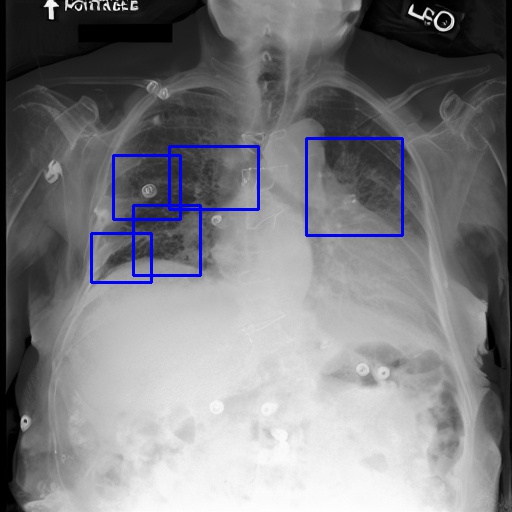} \\
            \includegraphics[width=\linewidth]{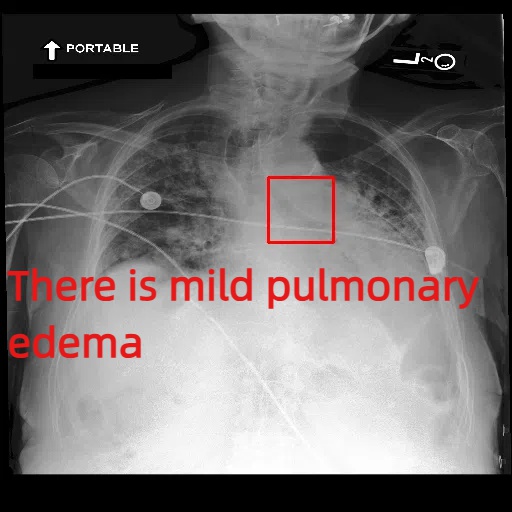} \\
            \includegraphics[width=\linewidth]{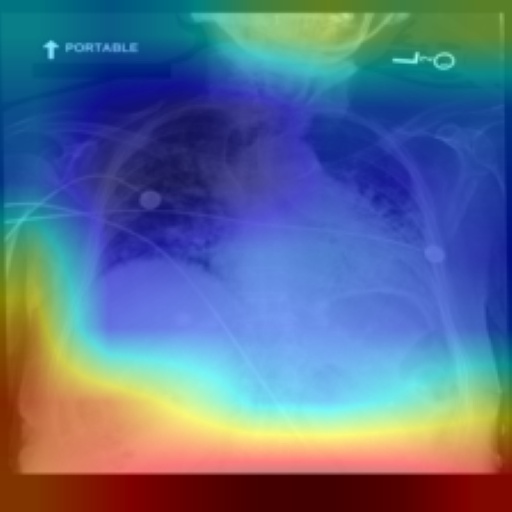} \\
            \makebox[\linewidth]{\centering \large \bl{`vascular congestion'}}
        \end{minipage} &
        \begin{minipage}{0.15\textwidth}
            \makebox[\linewidth]{\centering \large Lung Opacity} \\
            \includegraphics[width=\linewidth]{img/figure5/512_512resize/79p10439781_s51441976_3d0754cf-6b313d54-5c41bc32-9f042b6f-4f2f7051_init.png} \\
            \includegraphics[width=\linewidth]{img/figure5/512_512resize/079_remove_lung_opacity_box.png} \\
            \includegraphics[width=\linewidth]{img/figure5/fix_text/79_1_right_upper_lobe_opacities_are_unchanged_text.png} \\
            \includegraphics[width=\linewidth]{img/figure5/512_512resize/79_0042_opacities.png} \\
            \makebox[\linewidth]{\centering \large \bl{`opacities'}}
        \end{minipage} &
        \begin{minipage}{0.15\textwidth}
            \makebox[\linewidth]{\centering \large Consolidation} \\
            \includegraphics[width=\linewidth]{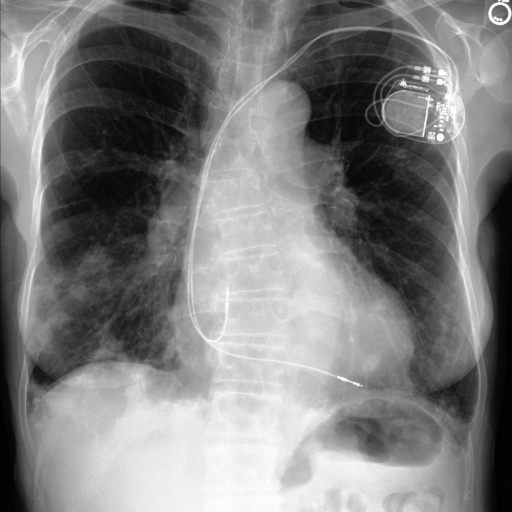} \\
            \includegraphics[width=\linewidth]{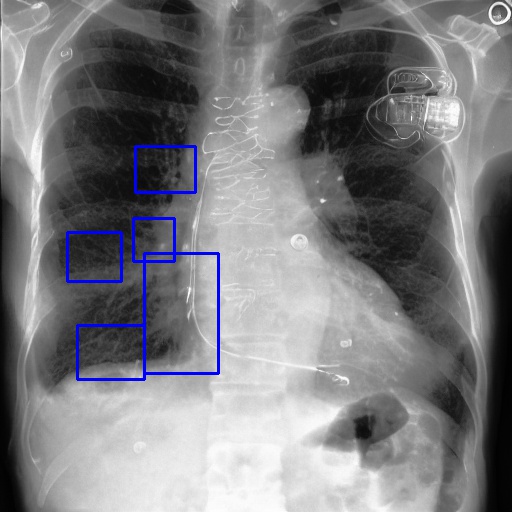} \\
            \includegraphics[width=\linewidth]{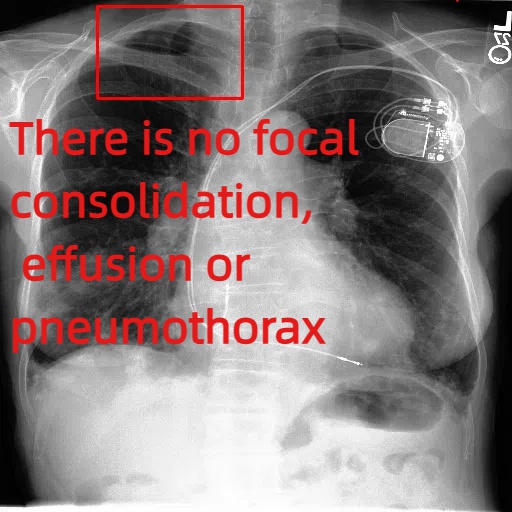} \\
            \includegraphics[width=\linewidth]{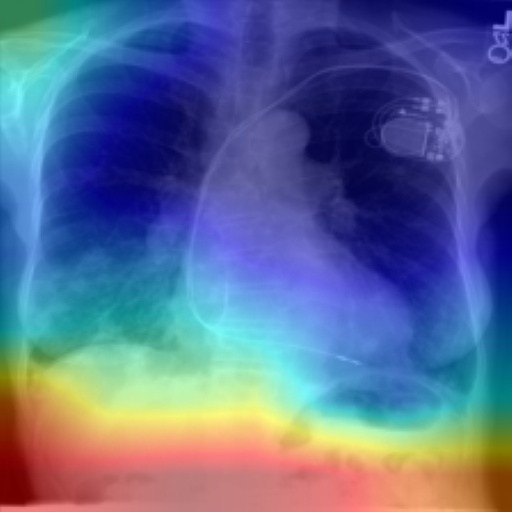} \\
            \makebox[\linewidth]{\centering \large \bl{`consolidation'}}
        \end{minipage}
    \end{tabular}}
    \caption{Qualitative comparison with cross attention and RGRG methods on the MIMIC-CXR dataset. 1st row is the initial images, 2nd row is the counterfactuals generated by CVLM method, 3rd row is the images with the bounding box and the text generated by RGRG method, and 4th row is the heatmap with the attention entities (the blue text) generated by the cross attention method. Note: The counterfactuals in the figure all achieve the cyclic manipulation in the regenerated report. In this image, CVLM and attention are model-agnostic explanation methods applied to the findings in the generated reports from R2GenCMN, which are confirmed to exist in the GT report. In contrast, RGRG are explainable report generator which generates its report separately based on specific regions.}
    \label{fig:seven_columns}
\end{figure*}

\begin{figure}[ht]
    \centering
    \begin{minipage}{0.23\textwidth}  
        \centering
        \includegraphics[width=0.49\linewidth]{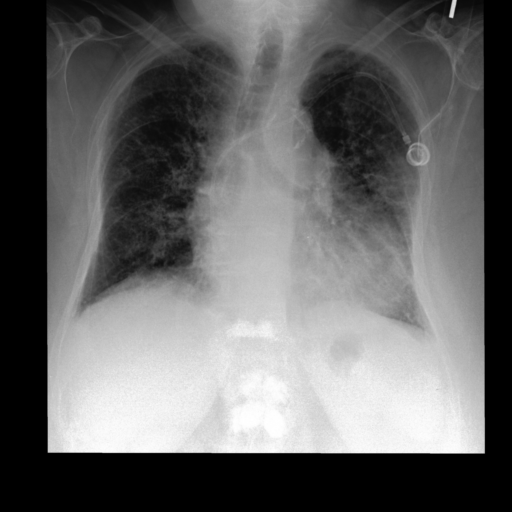} 
        \includegraphics[width=0.49\linewidth]{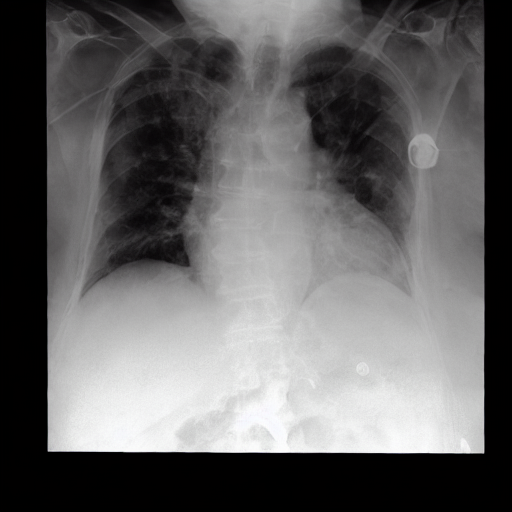} 
        \caption*{(a) Remove the ``cardiomegaly'' sentence from the raw report.}
    \end{minipage}
    \hfill 
    \begin{minipage}{0.23\textwidth}
        \centering
        \includegraphics[width=0.49\linewidth]{img/report_manipulation/89_init.jpg}
        \includegraphics[width=0.49\linewidth]{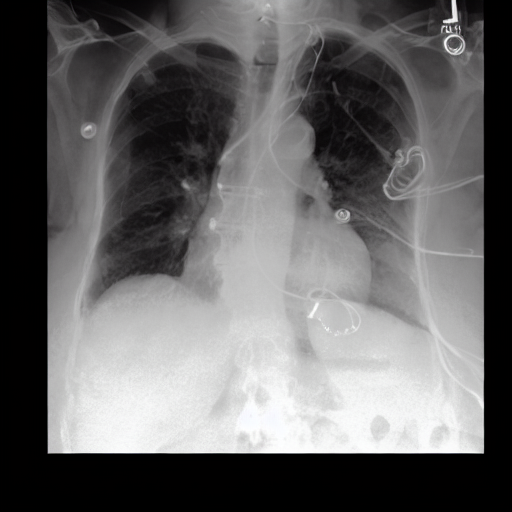} 
        \caption*{(b) Remove ``cardiomegaly'' from the reorganized report.}
    \end{minipage}
    \caption{Manipulation from stable diffusion trained with raw reports and cleaned reports, respectively. For both, the manipulation is removing the contents about the existence of cardiomegaly from the prompt. The organized prompt which focuses on the findings brings significant change while the cardiomegaly is removed.}
    \label{fig:manipulation}
\end{figure}
\subsubsection{Comparison of Different Report Generators}
We present the visual explanation results from the cyclic counterfactual X-rays in Fig.~\ref{Figure: differentmodels} for R2Gen and R2GenCMN respectively. Specifically, we remove the abnormalities from their generated reports and generate the counterfactuals respectively, and resend the counterfactual images to their respective report generators to see if the abnormalities have been removed in the generated report. 
For the query X-ray in Fig.~\ref{Figure: differentmodels}, R2GenCMN detected three abnormalities, while R2Gen detected two. Both models successfully removed the findings in their respective report generator outputs, and we can clearly observe the visual features contributing to the generated findings in their reports. 

As ``Atelectasis'' is the point of disagreement between the two models, we included the opinions of two radiologists to provide their judgment and identify the features indicative of its presence. We found that their assessments and the identified features closely matched the features detected by R2GenCMN. This suggests that R2GenCMN utilizes clinically recognized features for generating this finding, even though the ground-truth report did not include it. This also indicates that the explanations provided for the generated reports not only validate the findings of the models but also help identify potential missing findings in reports produced by humans. By presenting the generated report alongside its feature maps, our approach offers a more reliable and safer method for utilizing the generated reports in clinical practice. More cases are given in Appendix B\footnotemark[1].

\subsubsection{Benchmark against State-of-the-art Explanations}

Fig.\ref{fig:seven_columns} shows the different explanations generated for various abnormalities. Compared to the cross-attention method, our approach produces more accurate localization of the major findings. The heatmaps produced by the cross-attention method appear unstable; for instance, the findings for cardiomediastinal silhouette, opacity, vascular congestion, and consolidation in Fig.\ref{fig:seven_columns} are not localized within the lung areas.

The RGRG method provides reasonably interpretable results by associating findings with each anatomical region in its frame. However, upon examining the contents of its frame, it appears to have lower sensitivity to abnormalities, likely due to errors introduced by the pretrained detection model. While RGRG achieves internal interpretability, its framework cannot be adapted to other report generators that utilize different models or training datasets. In contrast, our proposed method offers precise localization explanations across various report generation models, as demonstrated in Fig.~\ref{Figure: differentmodels}.

Additionally, CVLM not only demonstrates the ability to identify fine-grained features within X-rays but also effectively manipulates features occurring in multiple locations, such as opacity. It can also detect large-scale global abnormalities, such as enlarged cardiomediastinum, significantly surpassing the feature identification capabilities of existing explanation methods.

\subsubsection{Ablation Study}

We compared the manipulation method within CVLM to a direct report manipulation approach, where Stable Diffusion is trained directly with unprocessed reports without pre-cleaning. The results in Fig. \ref{fig:manipulation} indicate that using organized prompts focused on findings results in significantly better performance compared to removing entire sentences from unstructured reports. For quantitative comparison, out of a total of 400 manipulations, the structured report manipulation achieved a much higher success rate (0.712) than the direct report manipulation approach (0.449) for the R2Gen model.

Finally, we analyzed the influence of the denoising step $T$ during diffusion model inference on achieving cyclic success. We evaluated $T$ values ranging from 20 to 100 and found that $T > 50$ yields the highest success rates. For efficiency, $T = 50$ was selected. The computation cost and numerical results are provided in Appendix B\footnotemark[1]. 
\footnotetext[1]{The appendix for this work is available at \url{https://arxiv.org/abs/2411.05261}.}

\subsection{Discussion}
\subsubsection{Broader Applications}
While there is an abundance of X-ray report generators, existing evaluation metrics primarily focus on assessing the accuracy of the generated language and the detected abnormalities. Our proposed method introduces a novel metric by enabling the verification of identified features, facilitating the comparison of image features used for the generated findings in the report. This provides a unique approach to evaluating the trustworthiness of the generated content from different report generators by examining whether they rely on clinically consistent image features. The proposed method is modality-agnostic and can be easily extended to report generators for other imaging modalities, such as MRI and CT scans, to improve the explainability of generated reports.

\subsubsection{Limitation and Future Work} 
Currently, our method relies on CheXbert, which classifies only 13 diseases for reorganizing the report. This limitation restricts CVLM to explaining these 13 specific keywords. However, given the success of large language models (LLMs), such as GPT-4 \cite{openai2023gpt4}, in accurately organizing reports, as demonstrated in a recent work \cite{zhang2024radgenome}, we can easily extend the proposed method to handle open vocabulary explanations by leveraging LLMs to extract a broader list of abnormalities from the generated reports, thereby enhancing the flexibility and applicability of CVLM.
Besides, we will include more human experts and return them the findings from different report generators and their corresponding visual contexts, to further benchmark the reliability of the existing report generators.

\section{Conclusion}

In this paper, we propose a counterfactual generation method for query X-rays input to the report generator. These counterfactual X-rays modify specific visual features, resulting in the disappearance of certain major findings in the originally generated reports. This approach provides users with insights into the underlying feature patterns utilized by the report generators.
Our method enhances the transparency of automated reports generated by existing report generators, serving as a valuable tool for experts to understand and evaluate the trustworthiness of AI-based reports. By improving interpretability and reliability, our approach strengthens confidence in the use of automated reports in real-world applications.

\section*{Acknowledgments}
This work is supported in part by the ERC IMI (101005122), the H2020 (952172), the MRC (MC/PC/21013), the Royal Society (IEC\textbackslash NSFC\textbackslash211235), the NVIDIA Academic Hardware Grant Program, the SABER project supported by Boehringer Ingelheim Ltd, NIHR Imperial Biomedical Research Centre (RDA01), Wellcome Leap Dynamic Resilience, UKRI guarantee funding for Horizon Europe MSCA Postdoctoral Fellowships (EP/Z002206/1), UKRI MRC Research Grant, TFS Research Grants (MR/U506710/1), and the UKRI Future Leaders Fellowship (MR/V023799/1).

\section*{Contribution Statement}
Yingying Fang and Zihao Jin (denoted by $^{*}$ on the first page) contributed equally to the conceptualization, methodology, investigation, and writing of this work. Shaojie Guo and Jinda Liu supported the benchmarking and visualization. Zhilin Yue contributed to the methodology and visualization. Yijian Gao, Junzhi Ning, and Zhili contributed to the review and editing of the manuscript. Simon Walsh provided clinical evaluation and review. Guang Yang (the corresponding author, denoted by $^{\dag}$ on the first page) contributed to the methodology, as well as the review and editing of the manuscript.

\newpage
\bibliographystyle{named}
\bibliography{ijcai25}
\newpage
\clearpage 
\phantomsection
\appendix
\renewcommand{\thesection}{\Alph{section}}

\onecolumn
\begin{center}
  \textbf{\Large Appendix of Cyclic Vision-Language Manipulator: Towards Reliable and Fine-Grained Image Interpretation for Automated  Report Generation}
\end{center}

\section{Unsupervised difference map generation}

The generation of counters to highlight the differences between the counterfactual and initial images, which subsequently alter the report generator's output, follows the pipeline outlined below:

\noindent\textbf{Step 1. Difference Map:} The absolute difference between the given initial image (1st column) and the counterfactual images, generated by removing the target findings from the reports (2nd column), is first calculated. A blur kernel of size 5x5 is then applied. To filter out noisy pixels, a threshold \(L\) is used, followed by the application of Otsu's method to calculate an adaptive threshold for each difference map, resulting in a binarised image (3nd column).

\noindent\textbf{Step 2. Extraction of Main Components} Morphological operations are employed to extract the major components from the separated pixels. Specifically, a morphological opening process is used to remove small objects with a fixed 3x3 kernel and an iteration count \(t1\). This is followed by a morphological dilation process, using the same kernel and an iteration count \(t2\), to connect nearby components. The iteration counts \(t1\) and \(t2\) are empirically determined and fixed for each object. The results of the opening and closing processes are displayed in the 4th and 5th columns, respectively.

\noindent\textbf{Step 3. Component Visualisation:} The extracted components are visualised by assigning each a distinct colour, as shown in the 6th column.

\noindent\textbf{Step 4. Component Filtering:} Components with areas smaller than 5\% of the total component area are removed. The top \(K\) components (with \(K=5\)) are selected as the final result. If the reserved components are fewer than the set threshold, all the components will be shown accordingly. For the identifiction related to the `Cardiomegaly' manipulation, we will apply one more step to remove the frames outside the heart areas by applying a heart mask.

The parameters \(L\), \(t1\), and \(t2\) are selected and fixed for each object based on empirical evidence. The parameter \(L\) ranges between 0 and 25 for images represented by integers between 0 and 255, \(t1\) ranges from 2 to 4, and \(t2\) ranges from 3 to 4. The specific parameters used for manipulating different findings are illustrated in the samples shown in Figure \ref{pipeline}.

\begin{figure}[!h]
    \centering
    \makebox[0.11\linewidth]{\footnotesize{\textbf{Init}}}
    \makebox[0.11\linewidth]{\footnotesize{Counterfactual}}
    \makebox[0.11\linewidth]{\footnotesize{\textbf{Difference}}}
    \makebox[0.11\linewidth]{\footnotesize{\textbf{Opening}}}
    \makebox[0.11\linewidth]{\footnotesize{\textbf{Dilation}}}
    \makebox[0.11\linewidth]{\footnotesize{\textbf{Components}}}
    \makebox[0.11\linewidth]{\footnotesize{\textbf{Top K}}}
    \makebox[0.11\linewidth]{\footnotesize{\textbf{Frame}}}
    \\
    \makebox[0.11\linewidth]{\footnotesize{Init}}
    \makebox[0.11\linewidth]{\footnotesize{`Effusion'}}
    \makebox[0.11\linewidth]{\footnotesize{L=25}}
    \makebox[0.11\linewidth]{\footnotesize{t1=3}}
    \makebox[0.11\linewidth]{\footnotesize{t2=4}} 
    \makebox[0.11\linewidth]{\footnotesize{}}
    \makebox[0.11\linewidth]{\footnotesize{K=1}}
    \makebox[0.11\linewidth]{\footnotesize{Frame}}
    \\
    \includegraphics[width=0.11\linewidth]{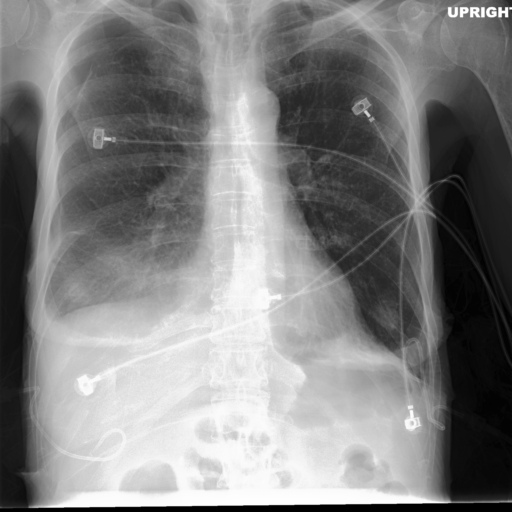} 
    \includegraphics[width=0.11\linewidth]{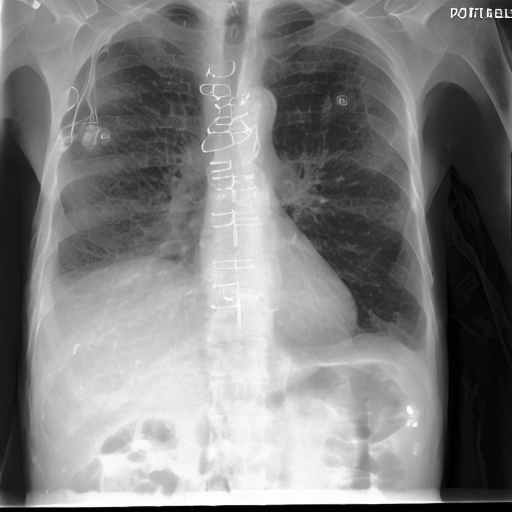} 
    \includegraphics[width=0.11\linewidth]{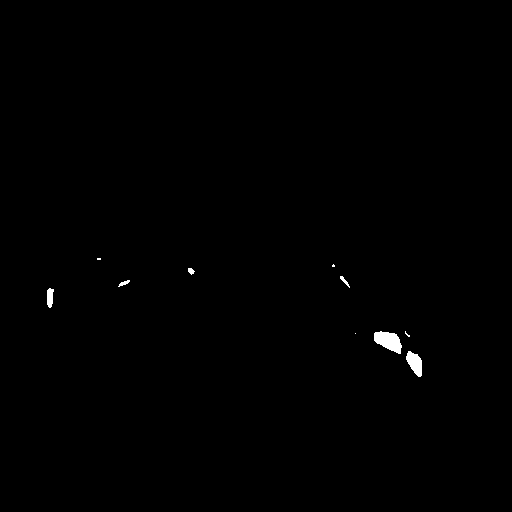} 
    \includegraphics[width=0.11\linewidth]{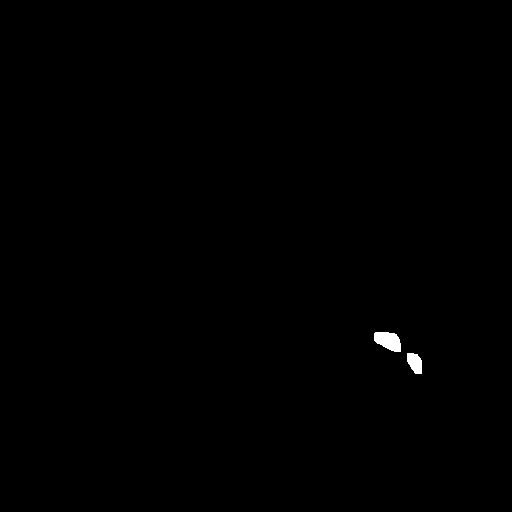} 
    \includegraphics[width=0.11\linewidth]{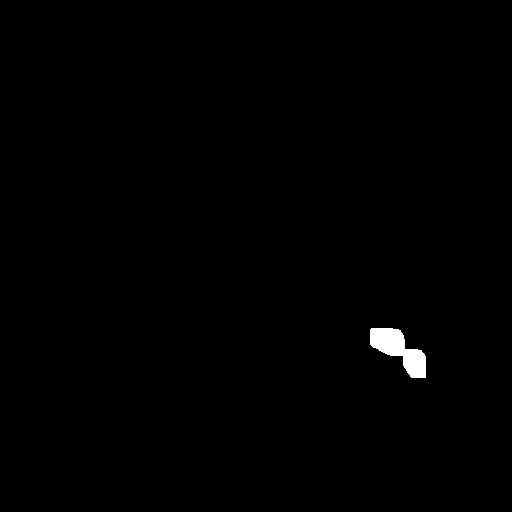} %
    \includegraphics[width=0.11\linewidth]{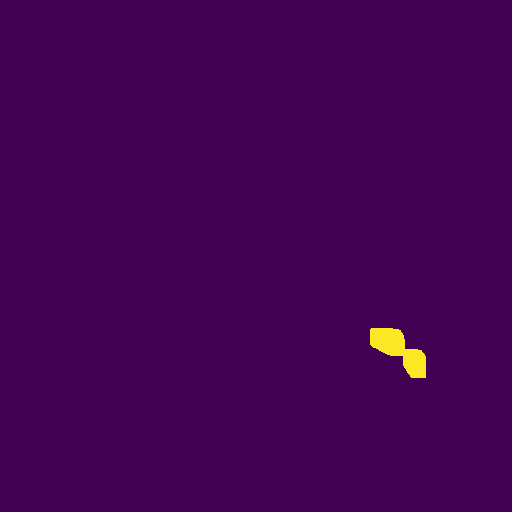} 
    \includegraphics[width=0.11\linewidth]{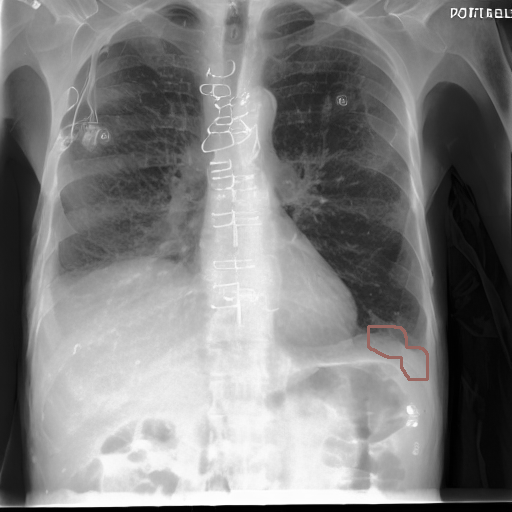}
    \includegraphics[width=0.11\linewidth]{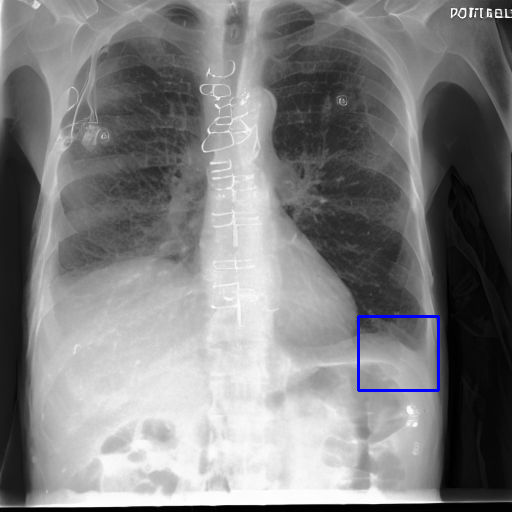} \\
    \centering{(a) Difference frames in the counterfactual images without `Pleural Effusion' in the report.}

    \centering
    \makebox[0.11\linewidth]{\footnotesize{Init}}
    \makebox[0.11\linewidth]{\footnotesize{`Enlarged'}}
    \makebox[0.11\linewidth]{\footnotesize{L=25}}
    \makebox[0.11\linewidth]{\footnotesize{t1=3}}
    \makebox[0.11\linewidth]{\footnotesize{t2=4}} 
    \makebox[0.11\linewidth]{\footnotesize{}}
    \makebox[0.11\linewidth]{\footnotesize{K=4}}
    \makebox[0.11\linewidth]{\footnotesize{Frame}}
    \\
    \includegraphics[width=0.11\linewidth]{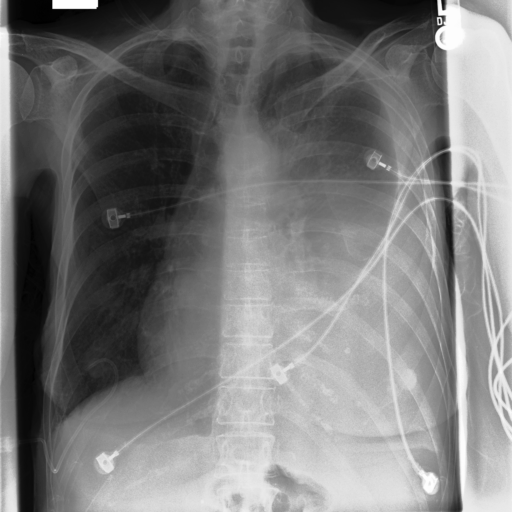} 
    \includegraphics[width=0.11\linewidth]{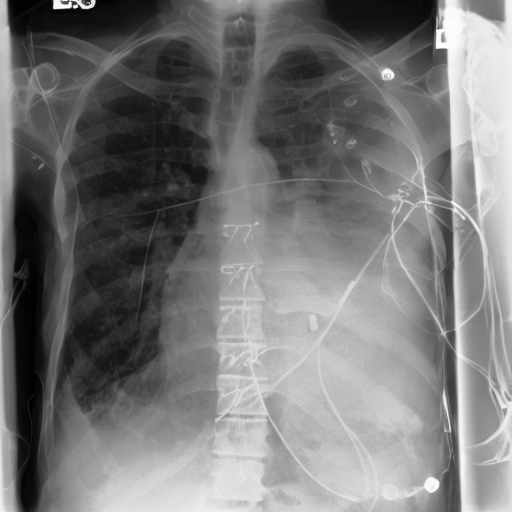} 
    \includegraphics[width=0.11\linewidth]{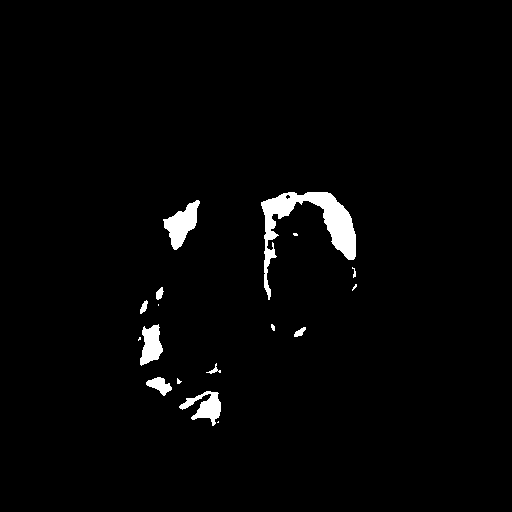} 
    \includegraphics[width=0.11\linewidth]{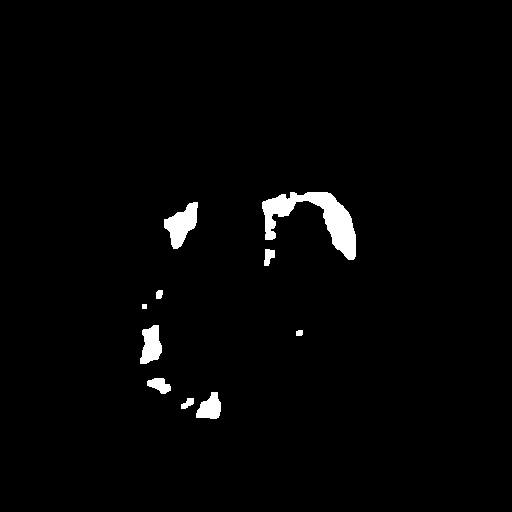} 
    \includegraphics[width=0.11\linewidth]{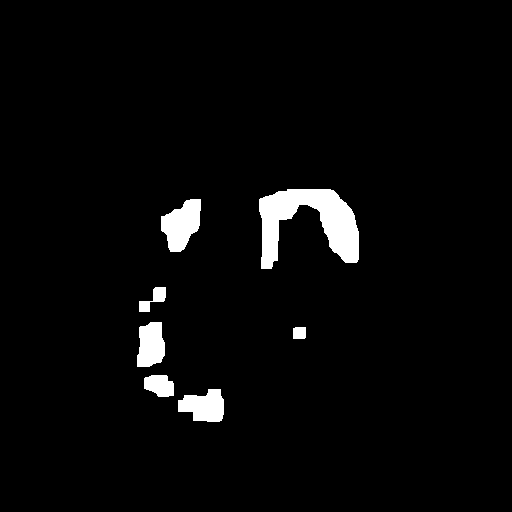} %
    \includegraphics[width=0.11\linewidth]{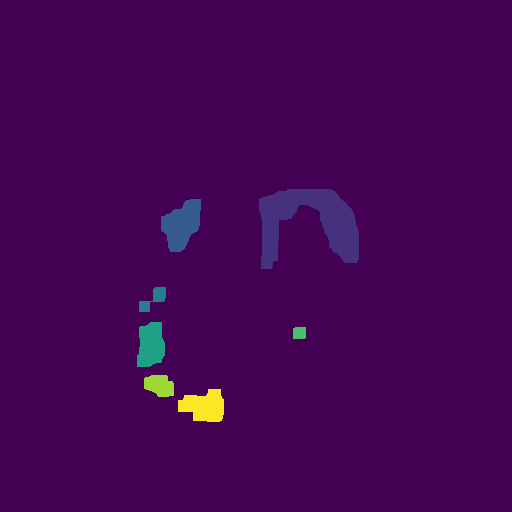} 
    \includegraphics[width=0.11\linewidth]{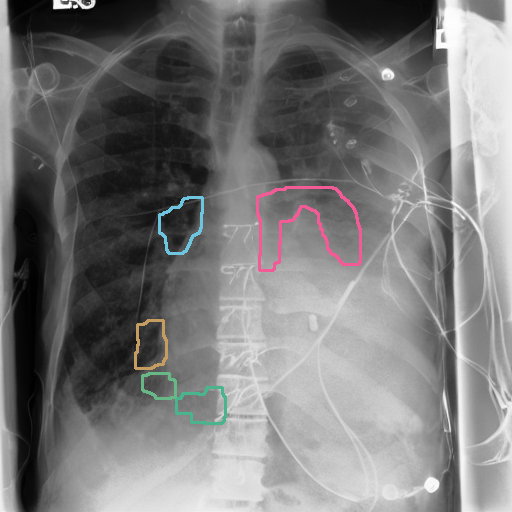}
    \includegraphics[width=0.11\linewidth]{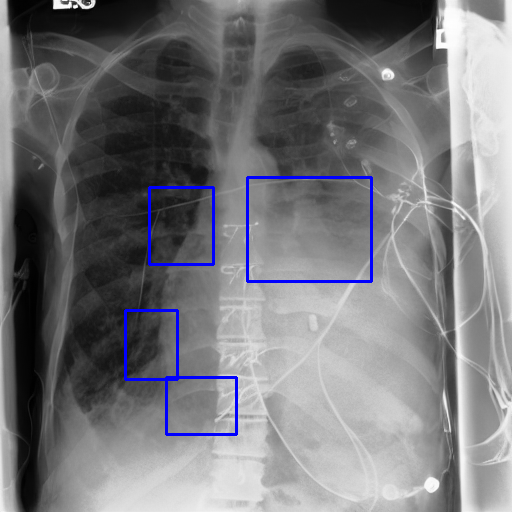} \\
    \centering{(b) Difference frames in the counterfactual images without `Enlarged Cardiomediastinum'.}

    \centering
    \makebox[0.11\linewidth]{\footnotesize{Init}}
    \makebox[0.11\linewidth]{\footnotesize{`Atelectasis'}}
    \makebox[0.11\linewidth]{\footnotesize{L=0}}
    \makebox[0.11\linewidth]{\footnotesize{t1=2}}
    \makebox[0.11\linewidth]{\footnotesize{t2=4}} 
    \makebox[0.11\linewidth]{\footnotesize{}}
    \makebox[0.11\linewidth]{\footnotesize{K=2}}
    \makebox[0.11\linewidth]{\footnotesize{Frame}}
    \\
    \includegraphics[width=0.11\linewidth]{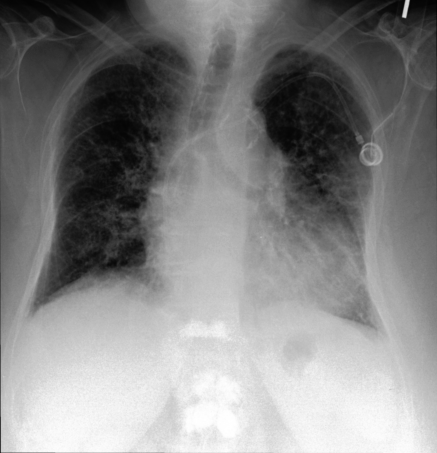} 
    \includegraphics[width=0.11\linewidth]{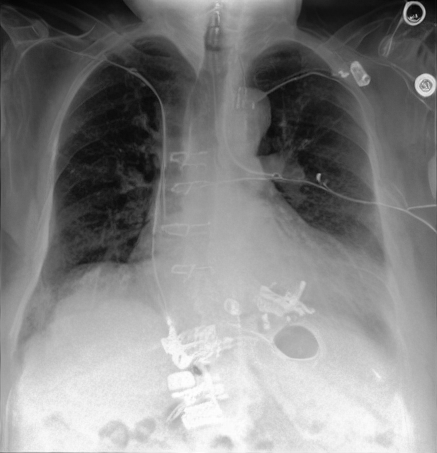}
    \includegraphics[width=0.11\linewidth]{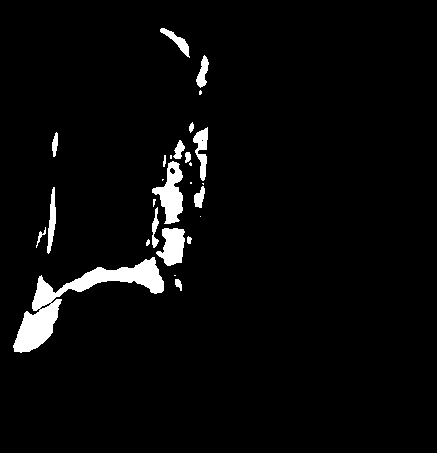} 
    \includegraphics[width=0.11\linewidth]{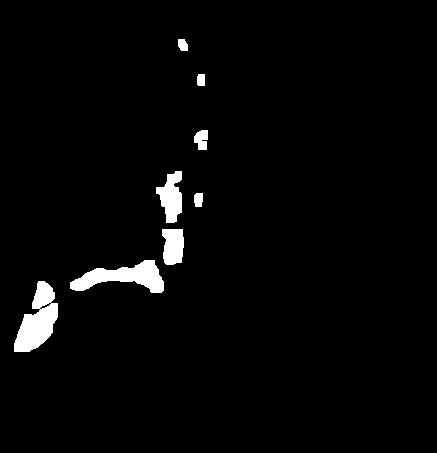} 
    \includegraphics[width=0.11\linewidth]{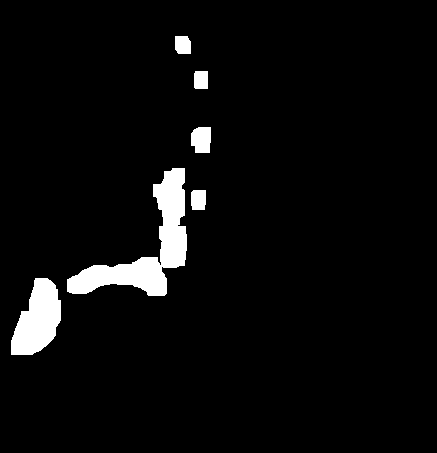} %
    \includegraphics[width=0.11\linewidth]{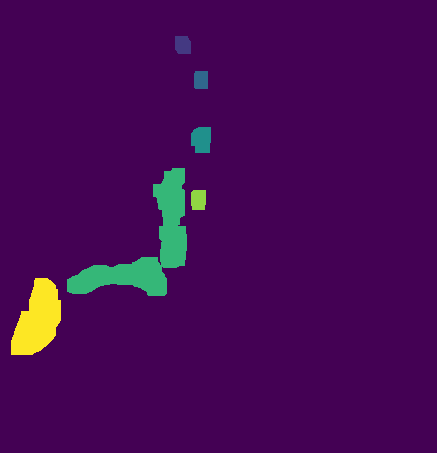} 
    \includegraphics[width=0.11\linewidth]{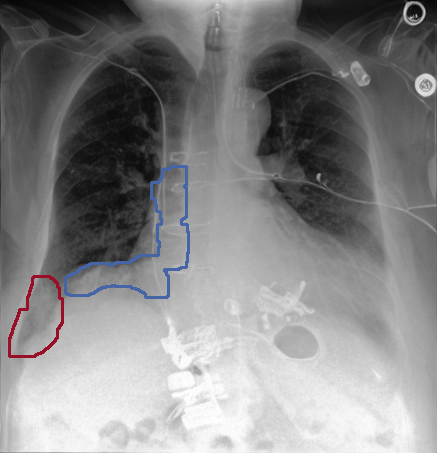}
    \includegraphics[width=0.11\linewidth]{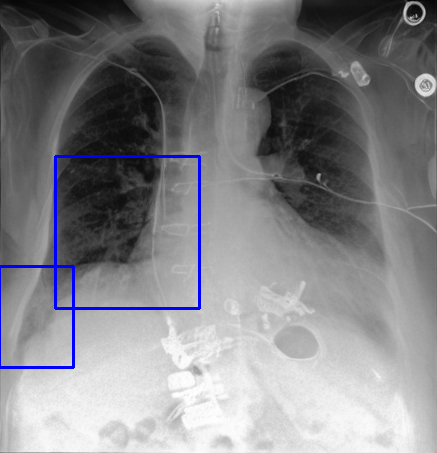} \\
    \centering{(c) Difference frames in the counterfactual images without `Atelectasis' in the report.}

    \centering
    \makebox[0.11\linewidth]{\footnotesize{Init}}
    \makebox[0.11\linewidth]{\footnotesize{`Enlarged'}}
    \makebox[0.11\linewidth]{\footnotesize{L=0}}
    \makebox[0.11\linewidth]{\footnotesize{t1=2}}
    \makebox[0.11\linewidth]{\footnotesize{t2=4}} 
    \makebox[0.11\linewidth]{\footnotesize{}}
    \makebox[0.11\linewidth]{\footnotesize{K=5}}
    \makebox[0.11\linewidth]{\footnotesize{Frame}}    
    \\
    \includegraphics[width=0.11\linewidth]{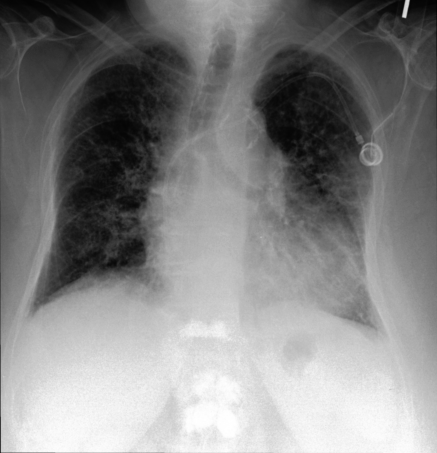} 
    \includegraphics[width=0.11\linewidth]{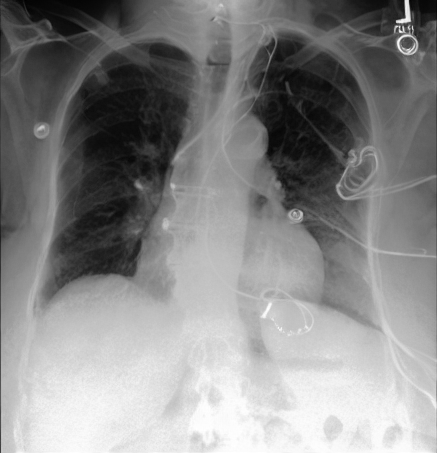} 
    \includegraphics[width=0.11\linewidth]{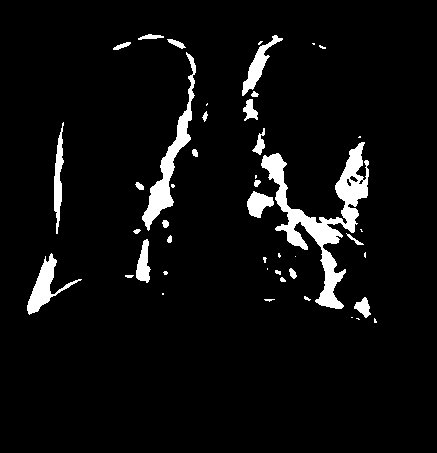} 
    \includegraphics[width=0.11\linewidth]{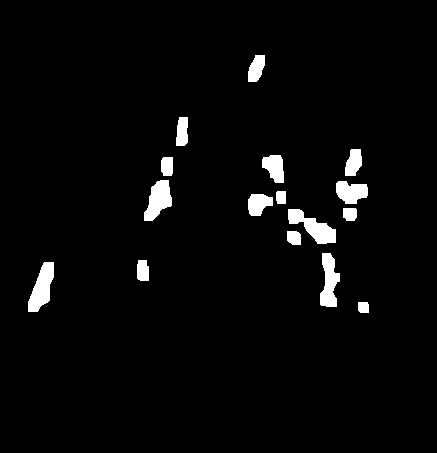} 
    \includegraphics[width=0.11\linewidth]{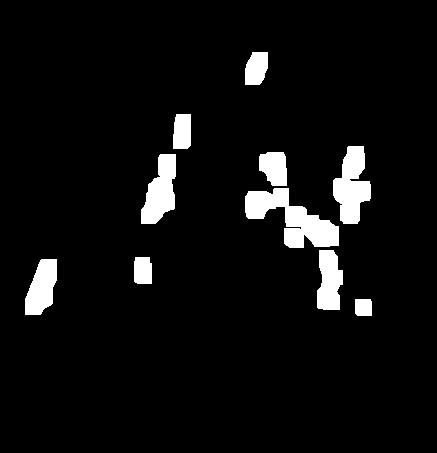} %
    \includegraphics[width=0.11\linewidth]{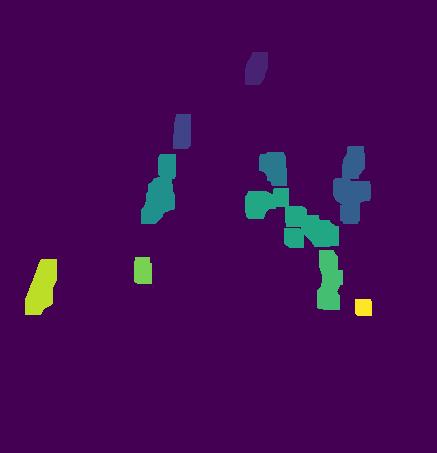} 
    \includegraphics[width=0.11\linewidth]{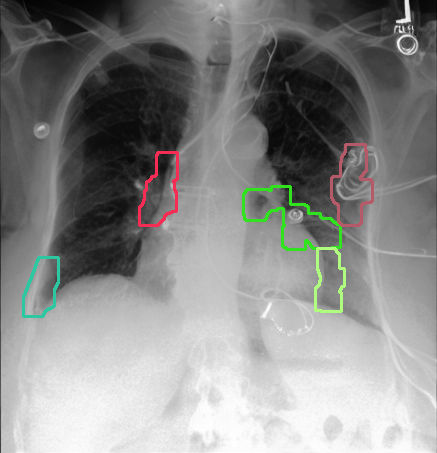}
    \includegraphics[width=0.11\linewidth]{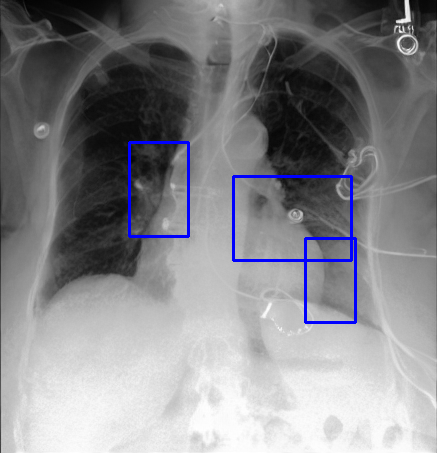}\\ 
    \centering{(d) Difference frames in the counterfactual images without `Cardiomegaly' in the report.}
        \end{figure}

\clearpage

\begin{figure}[!t]
    
    \centering
    \makebox[0.11\linewidth]{\footnotesize{Init}}
    \makebox[0.11\linewidth]{\footnotesize{`Device'}}
    \makebox[0.11\linewidth]{\footnotesize{L=15}}
    \makebox[0.11\linewidth]{\footnotesize{t1=2}}
    \makebox[0.11\linewidth]{\footnotesize{t2=4}} 
    \makebox[0.11\linewidth]{\footnotesize{}}
    \makebox[0.11\linewidth]{\footnotesize{K=1}}
    \makebox[0.11\linewidth]{\footnotesize{Frame}}    
    \\
    \includegraphics[width=0.11\linewidth]{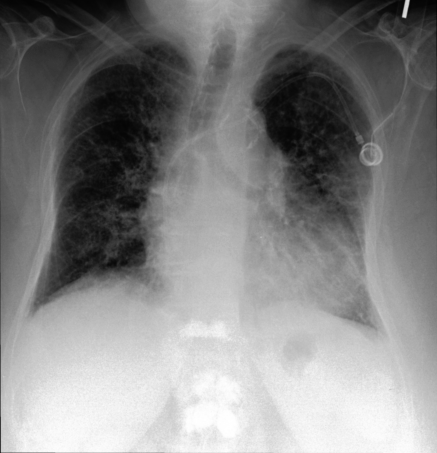} 
    \includegraphics[width=0.11\linewidth]{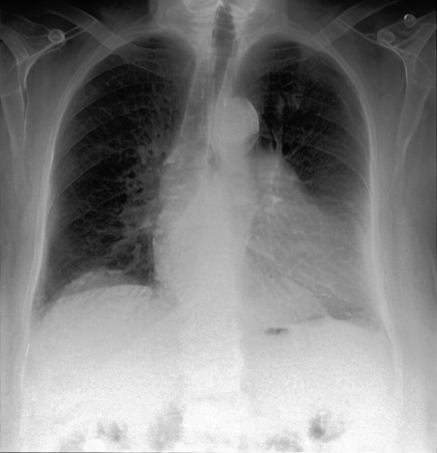} 
    \includegraphics[width=0.11\linewidth]{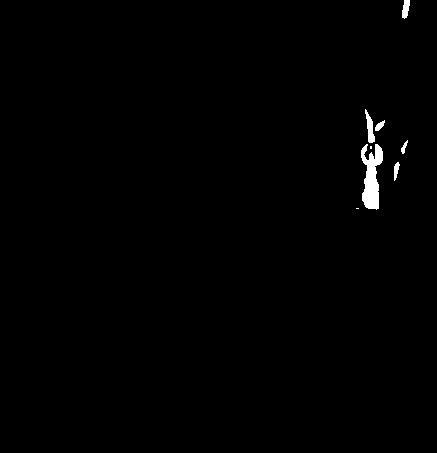} 
    \includegraphics[width=0.11\linewidth]{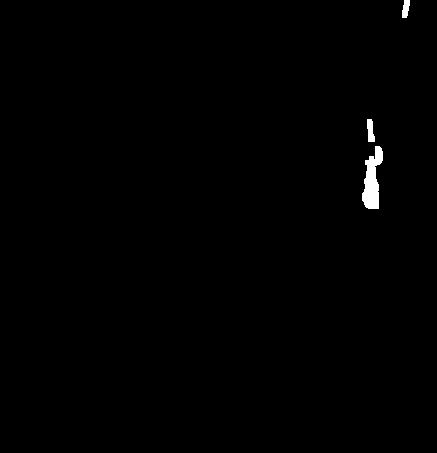} 
    \includegraphics[width=0.11\linewidth]{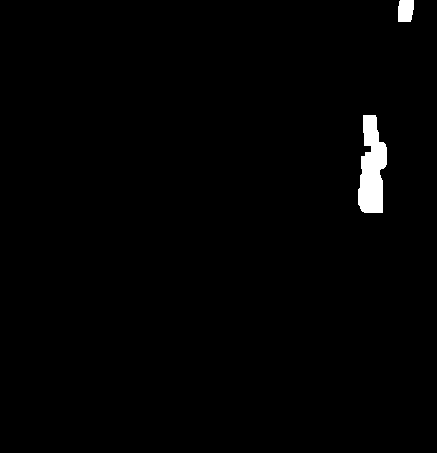} %
    \includegraphics[width=0.11\linewidth]{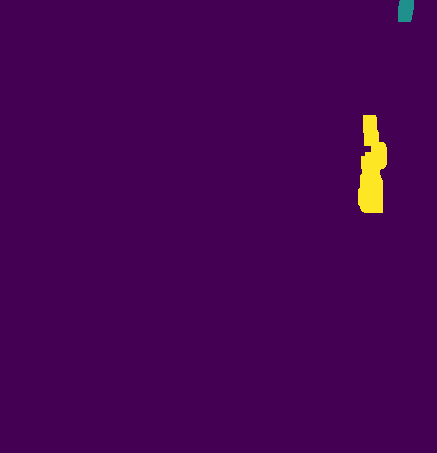} 
    \includegraphics[width=0.11\linewidth]{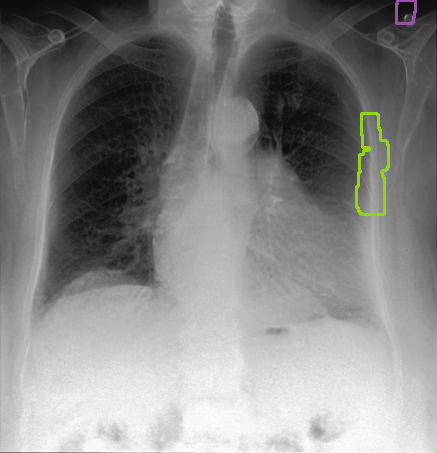}
    \includegraphics[width=0.11\linewidth]{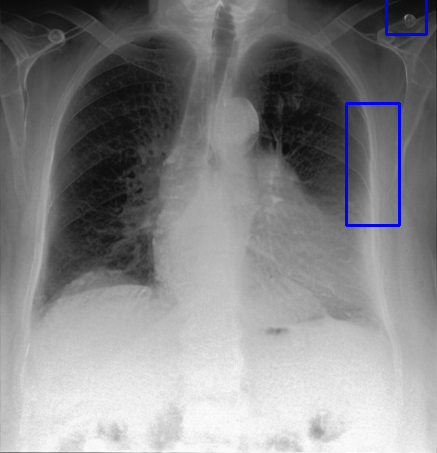}\\ 
    \centering{(e) Difference frames in the counterfactual images without `Support Devices' in the report.}\\
    \centering
    \centering
    \makebox[0.11\linewidth]{\footnotesize{Init}}
    \makebox[0.11\linewidth]{\footnotesize{`Consolidation'}}
    \makebox[0.11\linewidth]{\footnotesize{L=15}}
    \makebox[0.11\linewidth]{\footnotesize{t1=2}}
    \makebox[0.11\linewidth]{\footnotesize{t2=4}} 
    \makebox[0.11\linewidth]{\footnotesize{}}
    \makebox[0.11\linewidth]{\footnotesize{K=5}}
    \makebox[0.11\linewidth]{\footnotesize{Frame}}    
    \\
    \includegraphics[width=0.11\linewidth]{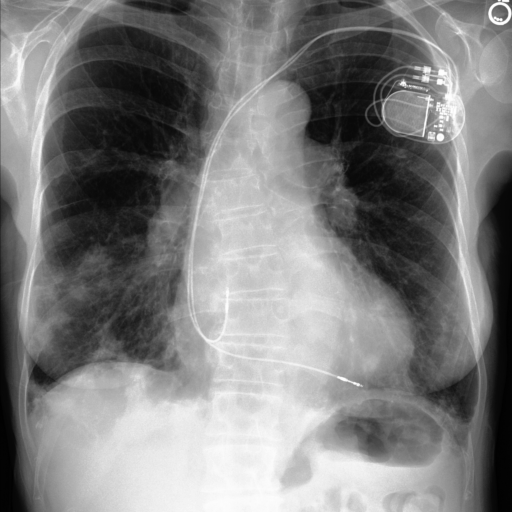} 
    \includegraphics[width=0.11\linewidth]{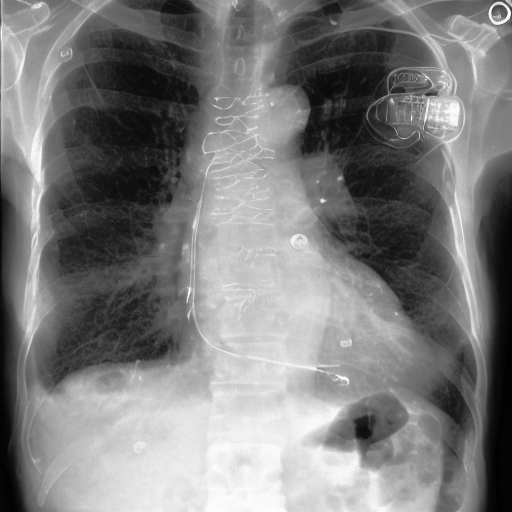}
    \includegraphics[width=0.11\linewidth]{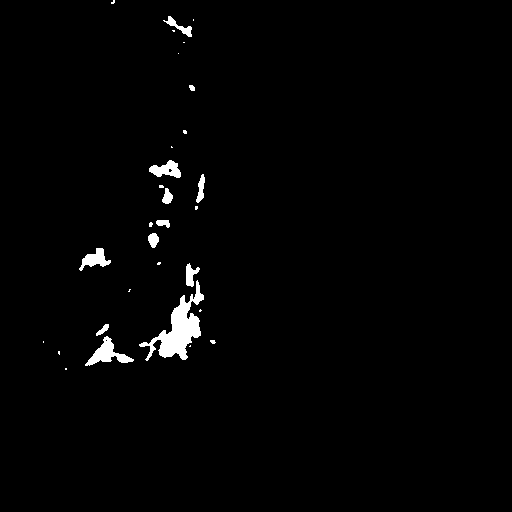} 
    \includegraphics[width=0.11\linewidth]{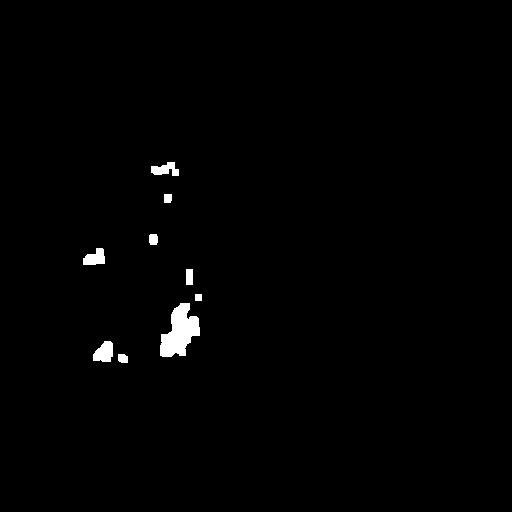} 
    \includegraphics[width=0.11\linewidth]{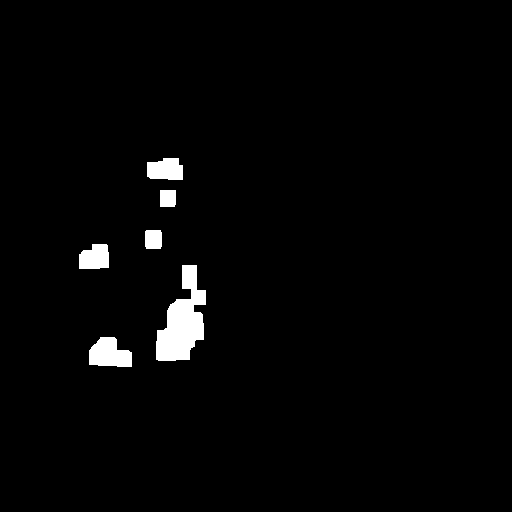} %
    \includegraphics[width=0.11\linewidth]{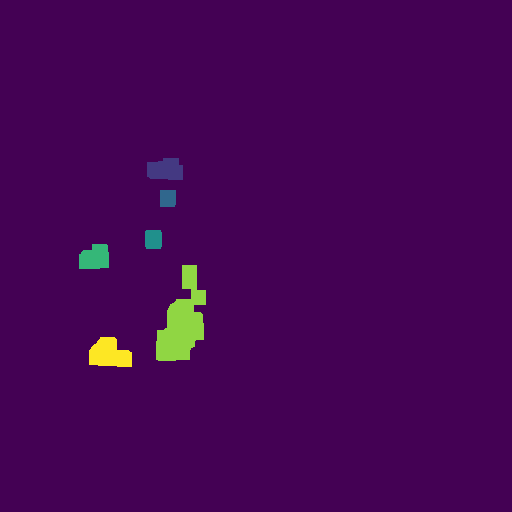} 
    \includegraphics[width=0.11\linewidth]{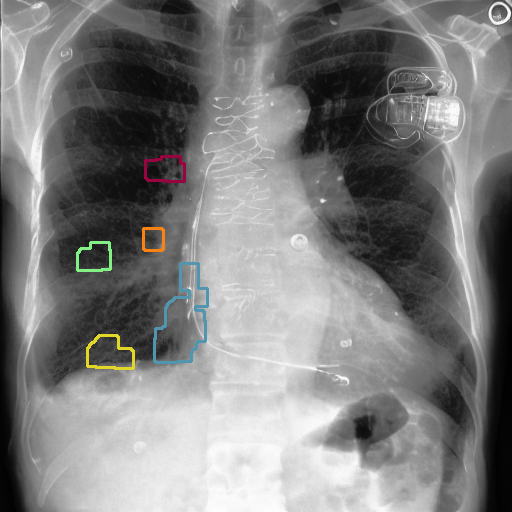}
    \includegraphics[width=0.11\linewidth]{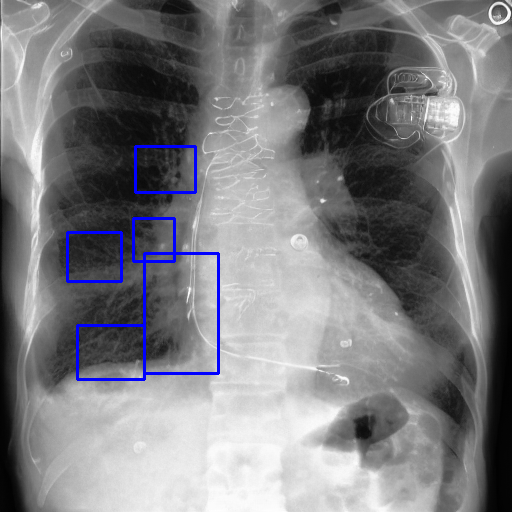} \\
    \centering{(f) Difference frames in the counterfactual images without `Consolidation' in the report.}

    \centering
    \makebox[0.11\linewidth]{\footnotesize{Init}}
    \makebox[0.11\linewidth]{\footnotesize{`Edema'}}
    \makebox[0.11\linewidth]{\footnotesize{L=15}}
    \makebox[0.11\linewidth]{\footnotesize{t1=2}}
    \makebox[0.11\linewidth]{\footnotesize{t2=4}} 
    \makebox[0.11\linewidth]{\footnotesize{}}
    \makebox[0.11\linewidth]{\footnotesize{K=5}}
    \makebox[0.11\linewidth]{\footnotesize{Frame}}    
    \\
    \includegraphics[width=0.11\linewidth]{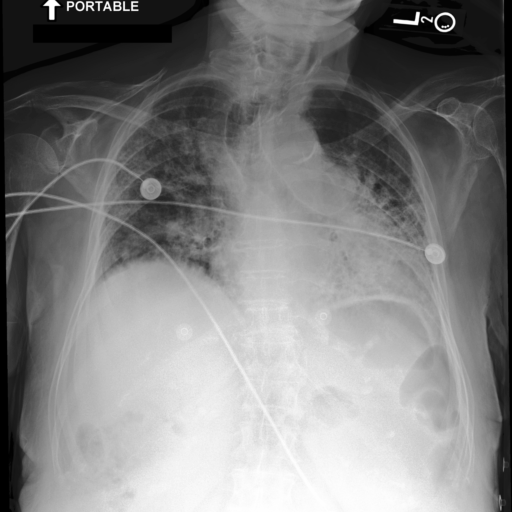} 
    \includegraphics[width=0.11\linewidth]{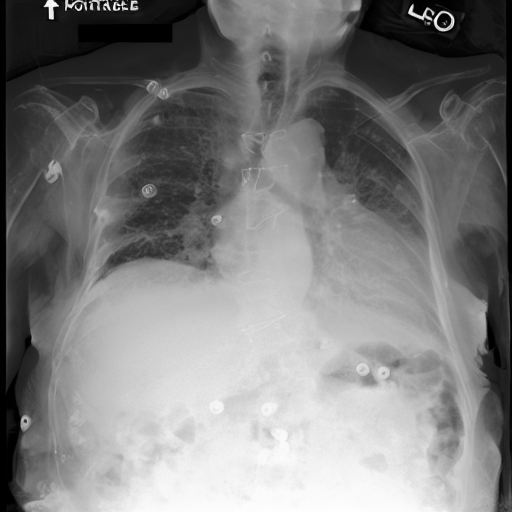}
    \includegraphics[width=0.11\linewidth]{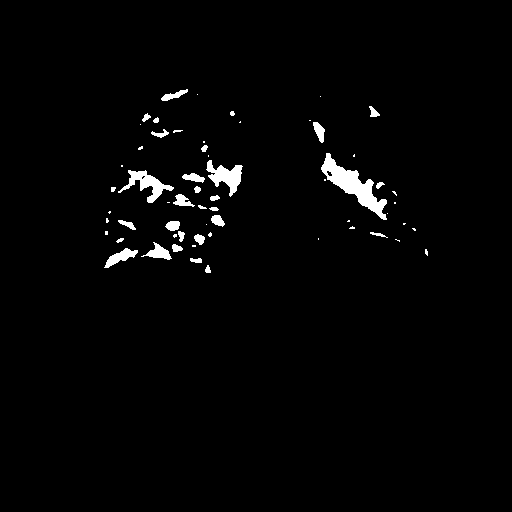} 
    \includegraphics[width=0.11\linewidth]{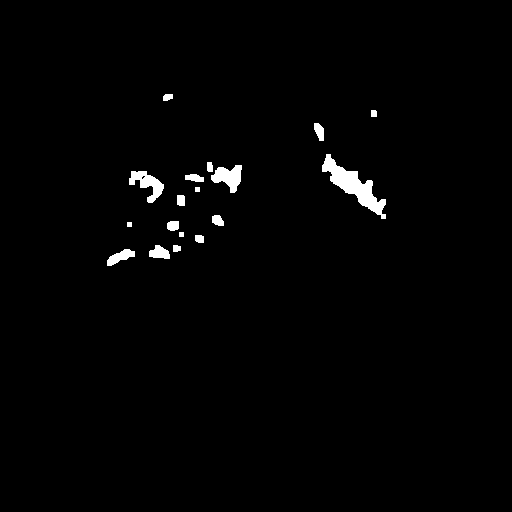} 
    \includegraphics[width=0.11\linewidth]{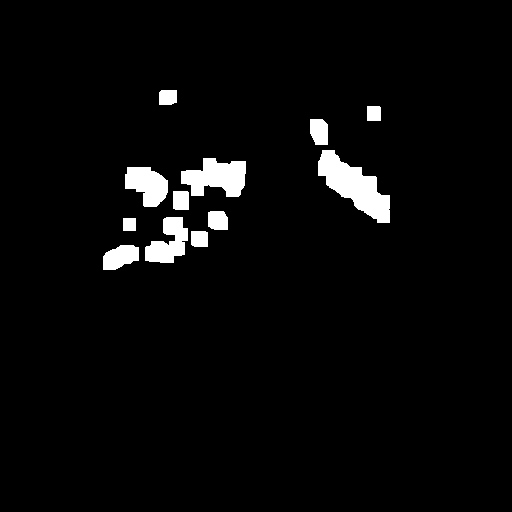} %
    \includegraphics[width=0.11\linewidth]{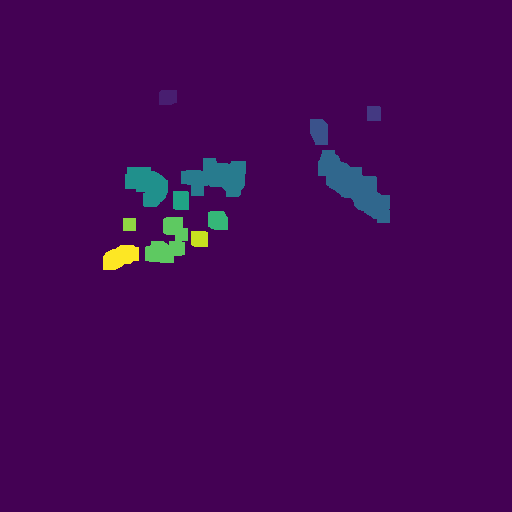} 
    \includegraphics[width=0.11\linewidth]{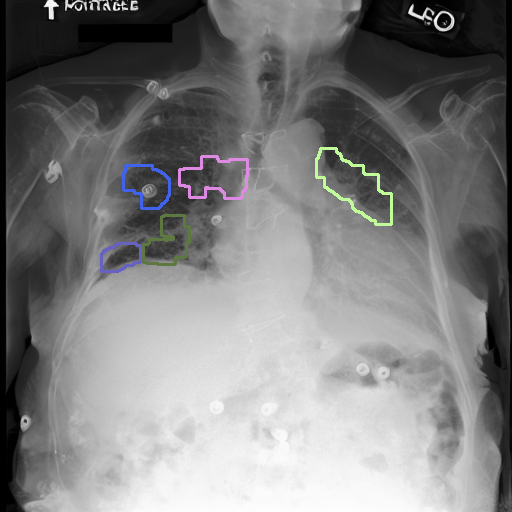}
    \includegraphics[width=0.11\linewidth]{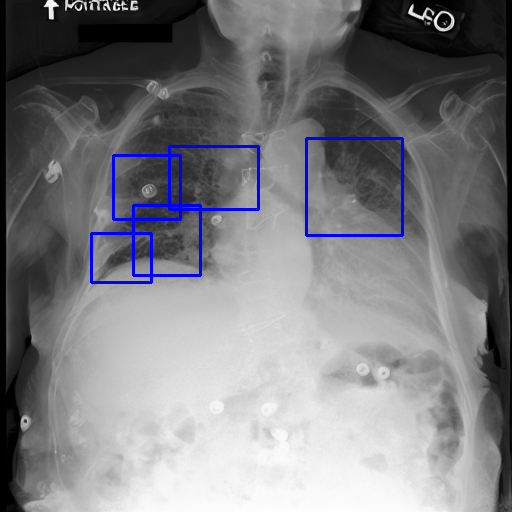} \\
    \centering{(g) Difference frames in the counterfactual images without `Edema' in the report.}
    \caption{Pipeline of generating the difference frames for different manipulation objects.}
    \label{pipeline}
\end{figure}

\begin{figure}[ht]

\end{figure}

\section{Results}

\subsection{Comparison of different report generators using CVLA}

\begin{figure}[!h]
  \centering
  \begin{subfigure}
    \centering
    \includegraphics[width=0.60\linewidth]{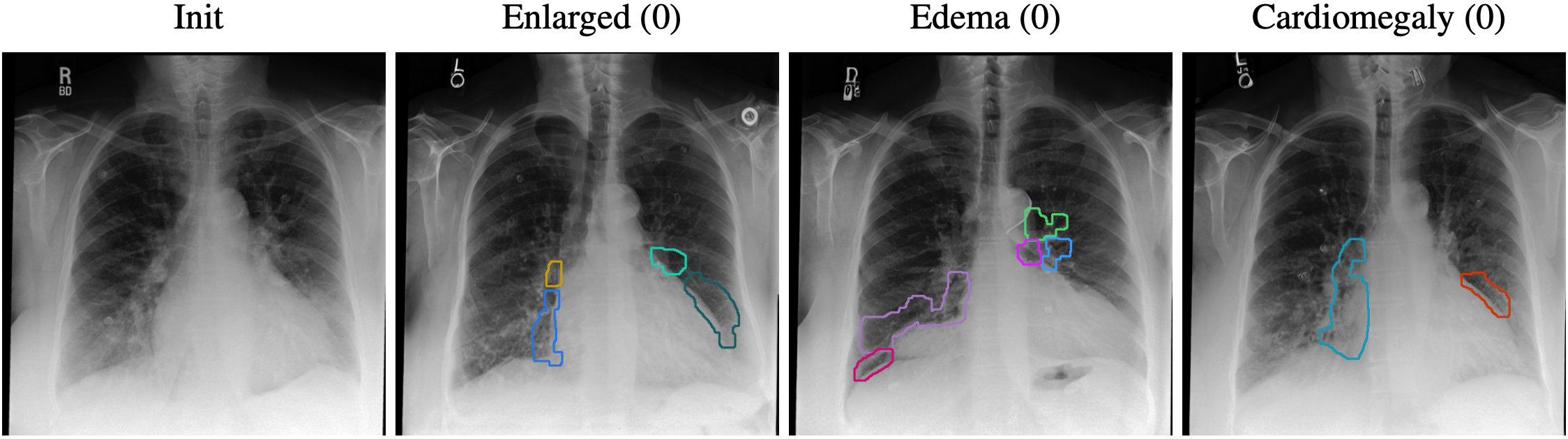}\\
    \centering{\small (a) Report findings of R2GenCMN:  Cardiomegaly, Enlarged Cardiomediastinum,  Edema}
  \end{subfigure}%
  \vspace{10pt}
  \begin{subfigure}
    \centering
    \includegraphics[width=0.60\linewidth]{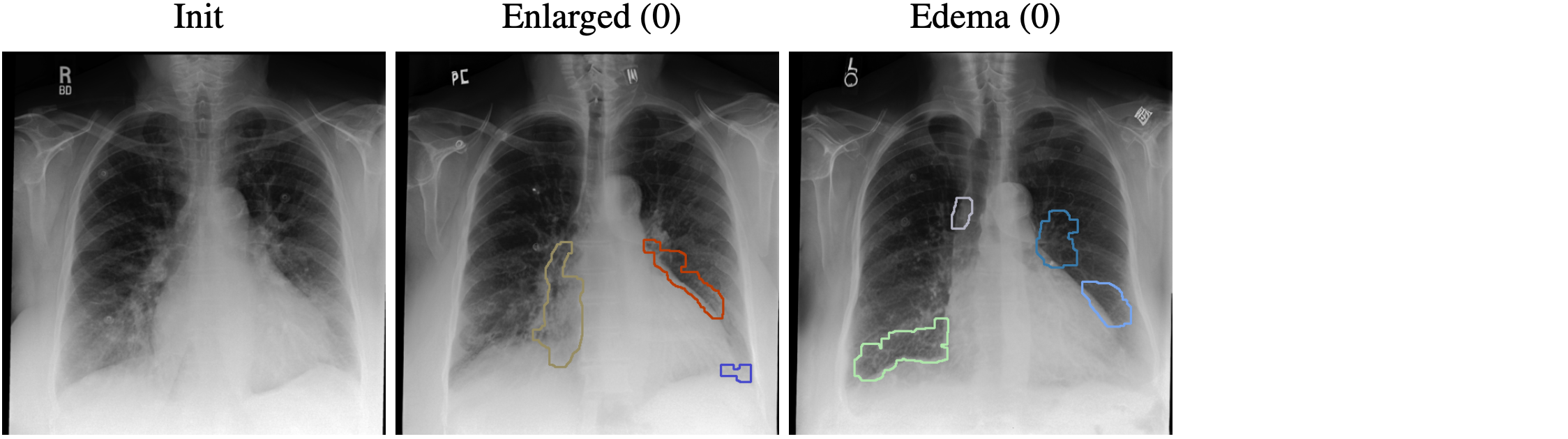}\\
    \centering{\small (b) Report findings of R2Gen:  Enlarged Cardiomediastinum,  Edema}
  \end{subfigure}

    \caption{Explanation results for the same query X-ray from different report generators. The counterfactual images are generated by the findings existing in the respective generated reports. (0) means the finding is removed in the regenerated report from the generated counterfactual images. }
    \label{Figure: differentmodels_278}
\end{figure}

\begin{figure}[!h]
  \centering
  \begin{subfigure}
    \centering
    \includegraphics[width=0.60\linewidth]{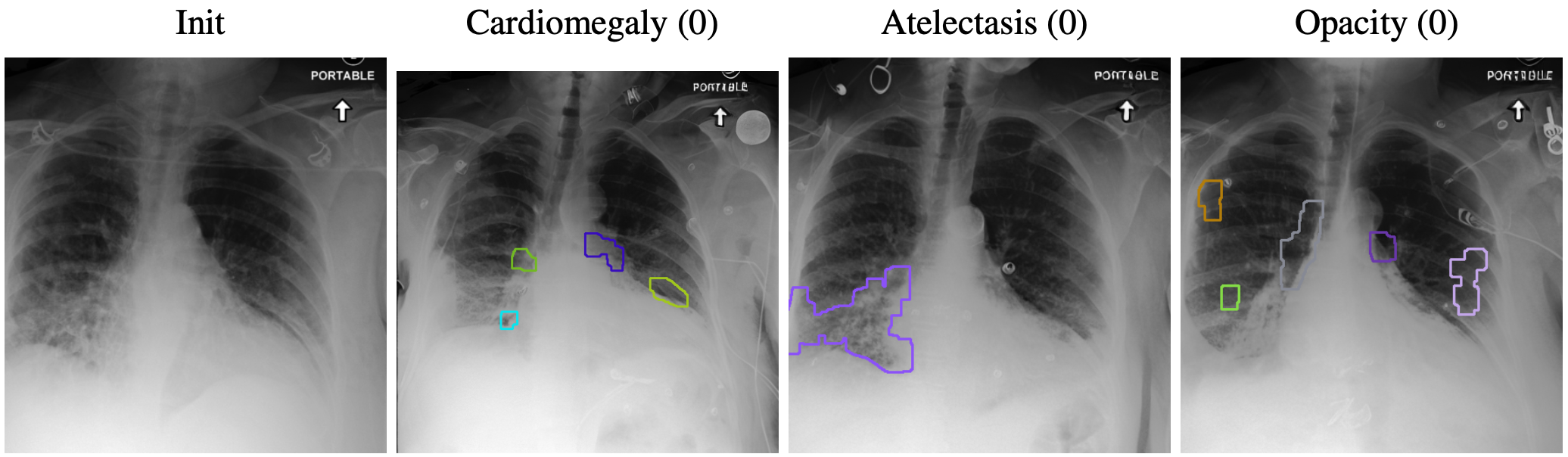}\\
    \centering{\small (a) Report findings of R2GenCMN:  Cardiomegaly, Atelectasis, Lung Opacity}
  \end{subfigure}%
  \vspace{10pt}
  \begin{subfigure}
    \centering
    \includegraphics[width=0.60\linewidth]{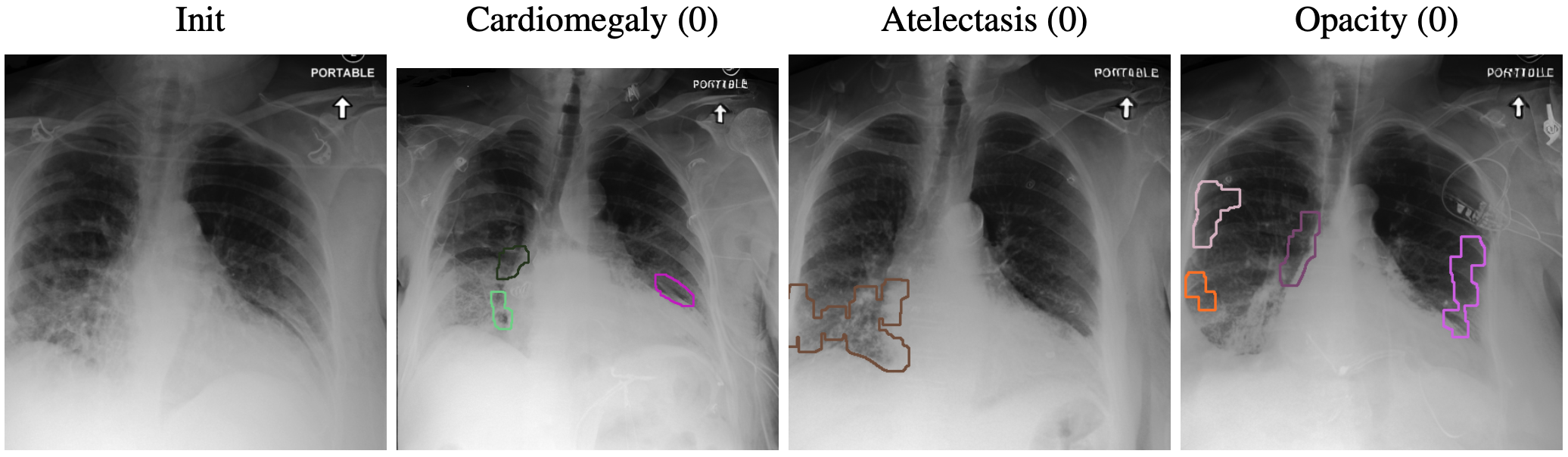}%
    \captionsetup{font={scriptsize}, labelfont={scriptsize}, skip=4pt}\\
    \centering{\small (b) Report findings of R2Gen:  Cardiomegaly, Atelectasis, Lung Opacity}
  \end{subfigure}
    \caption{Explanation results for the same query X-ray with different report generators. The counterfactual images are generated by the findings existing in the respective generated reports. (0) means the finding is removed in the regenerated report from the generated counterfactual images. }
    \label{Figure: differentmodels_422}
\end{figure}

In this section, we present supplementary examples using two query X-rays and the explanation results generated by two different report generators, R2GenCMN and R2GEN. The examples are illustrated in Figures \ref{Figure: differentmodels_278} and \ref{Figure: differentmodels_422}. The counterfactuals generated from these images remove the findings by feeding the regenerated reports back into the abnormality classification models. 

From these samples, we observe that different models can produce varying counterfactual images when given the same manipulation object. This variability assists in identifying the specific features that contribute to particular findings within each model. Moreover, by comparing the differing findings generated by the two models, we gain insights into the underlying reasons for these variations.

\subsection{Comparison of different interpretation methods}

In this section, we further analyse the explanation results by comparing the counterfactual images with their corresponding difference frames and cross-attention maps generated by the same model, R2GenCMN. Additionally, we compare these results with the reports and generated frames from the RGRG method. Our evaluation of the proposed method is based on the following observations:\\
\noindent \textbf{1. Localisation Correspondence:} We assess whether the localisation in the proposed frames derived from the counterfactual images aligns with the manipulated text generated by R2GenCMN.\\
\noindent \textbf{2. Cross-Attention Map Comparison:}
 We compare the cross-attention maps with the localisation provided by the proposed frames.\\
 \noindent \textbf{3. Localisation and Report Comparison:} We compare the localisation and corresponding reports from the frames generated by RGRG with those from the proposed method.\\
The supplementary results are presented in Figures 4 through 7, where bold, red, and blue fonts indicate the relevant statements in the different reports. The underlined are the words for generating the cross-attention maps. \\

\noindent \textbf{Case 1: Explaining the finding of `Device Support' in the generated reports}
\begin{figure}[!h]
    \centering
    \makebox[0.24\linewidth]{\footnotesize{Init}}
    \makebox[0.24\linewidth]{\footnotesize{RGRG}}
    \makebox[0.24\linewidth]{\footnotesize{Proposed}}
    \makebox[0.24\linewidth]{\footnotesize{`endotracheal tube'}}
    \\
    \includegraphics[width=0.24\linewidth]{img/figure5/512_512resize/307p11204646_s51866834_87c43c95-278ced86-fb0beb94-95ff11a9-8e8a8c3f_init.png} 
    \includegraphics[width=0.24\linewidth]{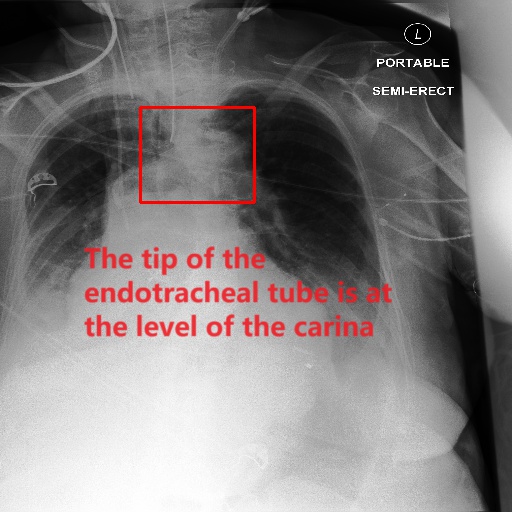}
    \includegraphics[width=0.24\linewidth]{img/figure5/512_512resize/307_remove_support_devices_box.png}
    \includegraphics[width=0.24\linewidth]{img/figure5/512_512resize/307_0002_endotracheal_tube.png}
    \caption{(Case 1) Explaining the finding of `Support Devices' in the generated reports}
    \label{opacity}
\end{figure}

\noindent\textbf{Human-labelled Report}: Comparison is made to previous study from \_\_\_.  The endotracheal tube and right-sided IJ central venous line are unchanged in position and appropriately sited.  \textbf{There is also a left-sided subclavian catheter with distal lead tip in the proximal SVC.}  There is stable cardiomegaly.  There are again seen bilateral pleural effusions and a left retrocardiac opacity.  There are no signs for overt pulmonary edema.  There are no pneumothoraces.\\
\noindent\textbf{Report of RGRG}:
There is no pneumothorax or pleural effusion. Right lower lobe atelectasis is unchanged. There is no evidence of pulmonary edema. Bibasilar atelectasis and pleural effusion are unchanged. Endotracheal tube is in standard position. As compared to the previous radiograph, the patient has received a nasogastric tube. \rd{The tip of the endotracheal tube is at the level of the carina.} Right internal jugular line tip is at the level of mid SVC. Moderate cardiomegaly. NG tube tip is in the stomach.\\
\noindent\textbf{Report of R2GenCMN}:
\bl{semi-upright portable radiograph of the chest demonstrates an \underline{endotracheal tube} ending 43 cm above the carina and an og tube courses into the stomach}. a right ij hemodialysis catheter ends in the mid svc . an enteric tube is in the esophagus with the tip out of field of view . lung volumes are low especially the lower lobes and the right upper lobe are chronically aerated . there is no large pleural effusion or pneumothorax . the cardiomediastinal and hilar contours are unchanged.\\

\noindent\textbf{Analysis:} The areas highlighted by the frames from the initial images align well with the reports generated by R2GenCMN. Compared to the cross-attention method, our approach demonstrates more precise localisation. In this scenario, the state-of-the-art method RGRG also provides a reasonable report for the indicated area, offering an explainable result.\\

\noindent \textbf{Case 2: Explaining the finding of `Lung Opacity' in the generated reports}
\begin{figure}[!h]
    \centering
    \makebox[0.24\linewidth]{\footnotesize{Init}}
    \makebox[0.24\linewidth]{\footnotesize{RGRG}}
    \makebox[0.24\linewidth]{\footnotesize{Proposed}}
    \makebox[0.24\linewidth]{\footnotesize{`opacities'}}\\
    \includegraphics[width=0.24\linewidth]{img/figure5/512_512resize/79p10439781_s51441976_3d0754cf-6b313d54-5c41bc32-9f042b6f-4f2f7051_init.png} 
    \includegraphics[width=0.24\linewidth]{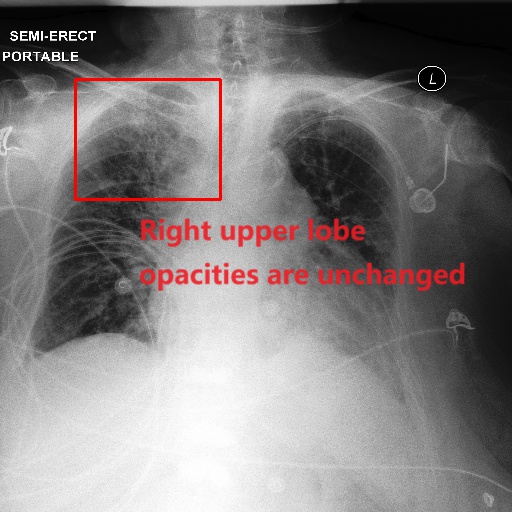}
    \includegraphics[width=0.24\linewidth]{img/figure5/512_512resize/079_remove_lung_opacity_box.png}
    \includegraphics[width=0.24\linewidth]{img/figure5/512_512resize/79_0042_opacities.png}
    \caption{(Case 2) Explaining the finding of `Lung Opacity' in the generated reports}
    \label{opacity}
\end{figure}

\noindent\textbf{Human-labelled Report}: 
In comparison with study of \_\_\_, there is little overall change. \textbf{Substantial cardiomegaly with bilateral opacifications most likely reflecting pulmonary edema.}  The possibility of supervening pneumonia would have to be raised in the appropriate clinical setting.  Central catheter remains in place.  Slight impression on the lower cervical trachea on the right could possibly represent a small thyroid mass. 
\\
\noindent\textbf{Report of RGRG:} \rd{Right upper lobe opacities are unchanged.} In comparison with the study of \_\_\_, \rd{there is increased opacity in the right upper lobe} and right lower lobe consistent with pulmonary edema. Bibasilar atelectasis is unchanged. There is mild pulmonary edema. There is no pneumothorax or pleural effusion. Moderate cardiomegaly and tortuosity of the aorta are unchanged. The mediastinal and hilar contours are unremarkable. Moderate cardiomegaly persists. NG tube tip is in the stomach.\\
\noindent\textbf{Report of R2GenCMN}: a left port-a-cath terminates in the mid svc unchanged . lung volume is low . cardiomediastinal silhouette is mostly unchanged compared to recent study . there is increased moderate pulmonary edema but overall has improved compared to upper-to-mid chest radiograph . \bl{patchy \underline{opacities} are increased from }. bilateral small pleural effusions likely present .
\\
\noindent\textbf{Analysis}:

The explanation results offer detailed localisation for the generated reports, which are more accurate than those produced by RGRG when compared to the human-annotated report.\\

\noindent \textbf{Case 3: Explaining the finding of `Edema' in the generated reports}\\
\begin{figure}[!h]
    \centering
    \makebox[0.24\linewidth]{\footnotesize{Init}}
    \makebox[0.24\linewidth]{\footnotesize{RGRG}}
    \makebox[0.24\linewidth]{\footnotesize{Proposed}}
    \makebox[0.24\linewidth]{\footnotesize{`vascular congestion'}}
    \\
    \includegraphics[width=0.24\linewidth]{img/figure5/512_512resize/143p10867202_s51723789_bcb5e90b-c7d3f928-7bd202ee-4e772a8f-e2240e90_init.png} 
    \includegraphics[width=0.24\linewidth]{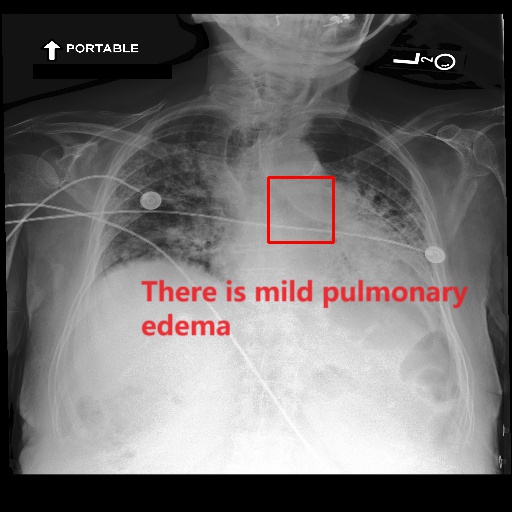}
    \includegraphics[width=0.24\linewidth]{img/figure5/512_512resize/143_remove_edema_box.png}
    \includegraphics[width=0.24\linewidth]{img/figure5/512_512resize/143_0006_vascular_congestion.png}
    \\
    \caption{(Case 3) Explaining the finding of `Edema' in the generated reports}
\end{figure}

\noindent\textbf{Human-labelled Report}: Lung volumes are low. Extensive bilateral opacities are unchanged from the prior examination and likely reflect the patient underlying severe interstitial lung disease.  \textbf{There is possibly increased opacification of the right lower lung, which may represent mild edema.} Hilar and cardiomediastinal contours are unchanged.  Calcification of the aortic arch is noted. There is no pneumothorax. There is no pleural effusion.\\
\noindent\textbf{Report of RGRG:} There is mild bibasilar atelectasis. \rd{There is mild pulmonary edema.} There is no pleural effusion or pneumothorax. There are no acute osseous abnormalities. The aorta is tortuous. The cardiomediastinal silhouette is unremarkable. Mediastinal contours are unremarkable. Moderate cardiomegaly persists. \\
\noindent\textbf{Report of R2GenCMN}: lung volumes are low . diffuse areas of parenchymal opacity are again noted raising concern for multifocal infection. \bl{there continues to be evidence of \underline{vascular congestion}}. cardiomediastinal silhouette is difficult to assess given low lung volumes and patient <unk> reticular opacities again seen . surgical clips are seen overlying the right neck and upper lung.\\
\noindent\textbf{Analysis}:

The explanation results offer detailed localization for the generated reports, which are more accurate than those produced by RGRG when compared to the human-annotated report.\\

\noindent \textbf{Case 4:  Explaining the finding of `Consolidation' in the generated reports}\\
\begin{figure}[!h]
    \centering
    \makebox[0.24\linewidth]{\footnotesize{Init}}
    \makebox[0.24\linewidth]{\footnotesize{RGRG}}
    \makebox[0.24\linewidth]{\footnotesize{Proposed}}
    \makebox[0.24\linewidth]{\footnotesize{`Consolidation'}}
    \\
    \includegraphics[width=0.24\linewidth]{img/figure5/512_512resize/132_init.png} 
    \includegraphics[width=0.24\linewidth]{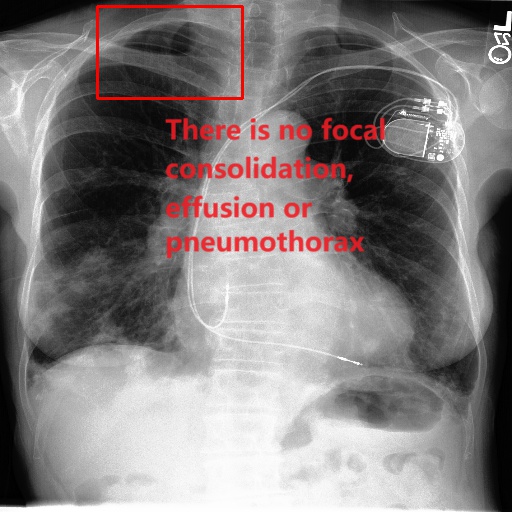}
    \includegraphics[width=0.24\linewidth]{img/figure5/512_512resize/132_remove_consolidation_box.png}
    \includegraphics[width=0.24\linewidth]{img/figure5/512_512resize/132_0013_consolidation.png}
    \caption{(Case 4) Explaining the finding of `Consolidation' in the generated reports}
    \label{Edema}
\end{figure}

\noindent\textbf{Human-labelled Report:} New multifocal parenchymal opacities in the lower and middle lobes bilaterally, which given concurrent increased hepatic density from \_\_\_ to \_\_\_, could represent amiodarone-induced pulmonary toxicity. Differential would includes infectious processes in the proper clinical setting or organizing pneumonia.  CT could be considered for further evaluation.  This was discussed with Dr \_\_\_ at noon by Dr \_\_\_ on \_\_\_ via phone.\\
\noindent\textbf{Report of RGRG:} There is no evidence of acute cardiopulmonary process. Right lower lobe pneumonia is unchanged. The mediastinal and hilar contours are normal. \rd{There is no focal consolidation, effusion, or pneumothorax. Bibasilar atelectasis is unchanged.} There are no acute osseous abnormalities. The cardiomediastinal silhouette is within normal limits. Moderate cardiomegaly is unchanged.\\
\noindent\textbf{Report of R2GenCMN:} a dual-lead left-sided pacemaker is again seen with leads extending to the expected positions of the right atrium and right ventricle . there are multifocal patchy opacities in the bilateral lung bases which on second chest ct were more sensitive for parenchymal abnormality on the prior ct . \bl{slight focal opacity in the right mid hemi thorax may be artifactual however underlying \underline{consolidation} is not excluded in the appropriate clinical setting}. the cardiac silhouette is not enlarged . there is mild gaseous distention of colon . mildly dilated stomach is seen not well assessed on the current study as \\
\noindent\textbf{Analysis:} Although certain findings are only detected by R2GenCMN and are not mentioned in the ground truth or RGRG reports, the explanation results from the proposed method offer a reasonable justification for these generated findings. This is valuable for human assessment of the reliability of the generated reports.\\

\noindent \textbf{Case 5: Analysis of Cases with Generated Milder Findings}\\

\begin{figure}[ht]
    \centering
    \begin{minipage}{0.45\textwidth}  
        \centering
        \makebox[0.3\linewidth]{\footnotesize{I}}
        \makebox[0.3\linewidth]{\footnotesize{M}}

        \includegraphics[width=0.3\linewidth]{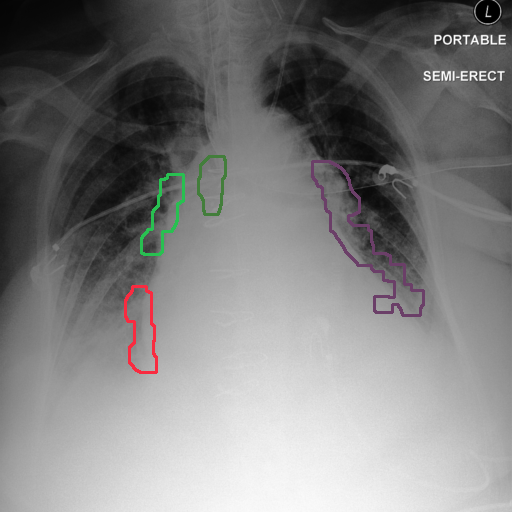} 
        \includegraphics[width=0.3\linewidth]{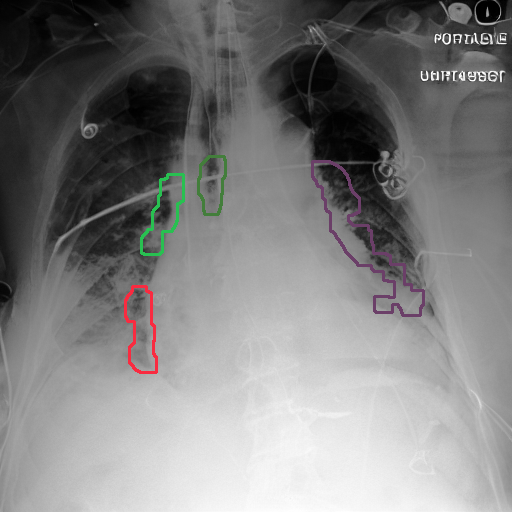} %
        \caption*{(a) case of `mild'}  
    \end{minipage}
    \hfill  
    \begin{minipage}{0.45\textwidth} 
    \centering
    \makebox[0.3\linewidth]{\footnotesize{I}}
    \makebox[0.3\linewidth]{\footnotesize{M}}\\  
    \includegraphics[width=0.3\linewidth]{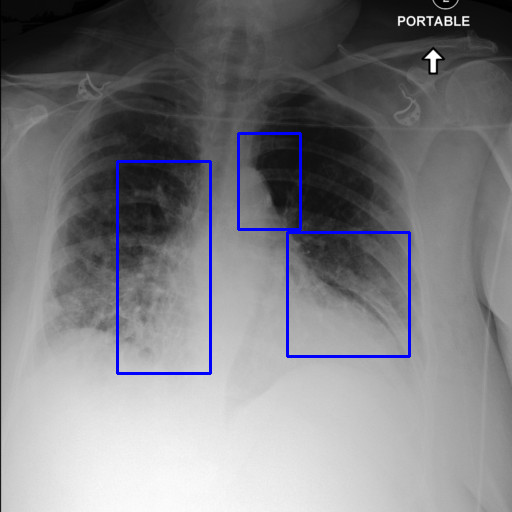}
    \includegraphics[width=0.3\linewidth]{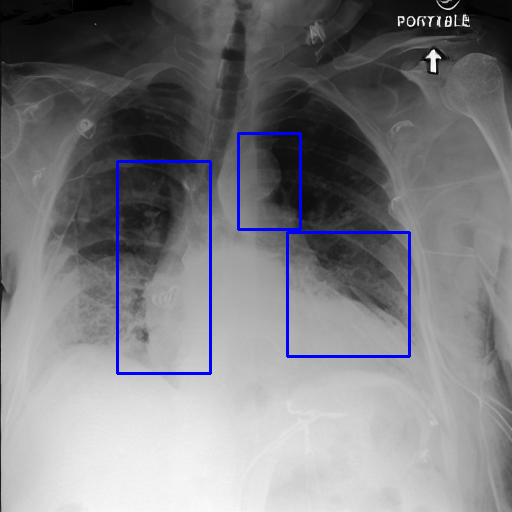}  
    \caption*{(b) case of `borderline'}
    \end{minipage}
    \caption{
    Explanations for varying extents of cardiomegaly (initial image (I) and manipulated result (M)))}
    \label{fig:extent}
\end{figure}

We further investigate CVLM‘s ability in explaining the subtle findings in the generated report, such as the findings with adjective  `mild' and `borderline' abnormalities. In Fig.~\ref{fig:extent}, we show two cases of detected `mild' and `borderline' cardiomegaly in the  report generated by R2GENCMN. The results show that CVLM can localize subtle features associated with generated findings
of minor extent.

\subsection{Ablation study in terms of inference times}
We analyzed the influence of the denoising step $T$ during diffusion model inference on achieving cyclic success. We evaluated $T$ values ranging from 20 to 100 and found that $T > 50$ yields the highest success rates (SRs)in  Table \ref{tab:denoising_steps}. We also compared the inference time of the proposed method with RGRG and cross attention methods in  Table \ref{tab:time}.
\vspace{10pt}

\begin{table}[ht]
\caption{Ablation study in terms of inference times on R2GenCMN}
\centering
\begin{tabular}{ccccccc}
\toprule
\textbf{Denoising steps: T} & 25 & 30 & 40 & 50 & 75 & 100 \\ \midrule
\textbf{Success Rate}                & 0.594 & 0.623 & 0.658 & 0.690 & 0.612 & 0.613  \\ \midrule
\textbf{Time (s)} & 2.86 & 3.85 & 4.26 & 5.16 & 7.26 & 9.61 \\ \bottomrule
\end{tabular}
\label{tab:denoising_steps}
\end{table}

\begin{table}[ht]
\caption{Comparison between different explanation method in terms of inference times}
\centering
\begin{tabular}{cccc}
\toprule
\textbf{Methods} & CVLM (Ours) & RGRG & Cross Attention \\ \midrule

\textbf{Time (s)} & 5.16 & 4.50 & 5.00 \\ \bottomrule
\end{tabular}
\label{tab:time}
\end{table}

\end{document}